\documentclass{article}

\PassOptionsToPackage{numbers, compress}{natbib}

\usepackage[final]{neurips_2022}

\usepackage[utf8]{inputenc} 
\usepackage[T1]{fontenc}    
\usepackage{hyperref}       
\usepackage{url}            
\usepackage{booktabs}       
\usepackage{amsfonts}       
\usepackage{nicefrac}       
\usepackage{microtype}      
\usepackage{xcolor}         

\usepackage{graphicx}
\usepackage{doi}
\usepackage{caption}
\usepackage{subcaption}
\usepackage{amsmath}
\usepackage{amsthm}
\usepackage{amssymb}
\usepackage{bbm}
\usepackage{tikz}
\usepackage{wrapfig}
\usepackage{comment}

\usepackage[bb=boondox]{mathalfa}

\newcommand{\reb}[1]{\textcolor{black}{#1}}

\title{Concept Embedding Models:\\Beyond the Accuracy-Explainability Trade-Off}

\author{%
  Mateo Espinosa Zarlenga\thanks{Equal contribution} \\
  University of Cambridge\\
  \texttt{me466@cam.ac.uk} \\
  \And
  Pietro Barbiero\footnotemark[1] \\
  University of Cambridge\\
  \texttt{pb737@cam.ac.uk} \\
  \And
  Gabriele Ciravegna \\
  Université Côte d’Azur, Inria, CNRS,\\ I3S, Maasai, Nice, France\\
  \texttt{gabriele.ciravegna@inria.fr} \\
  \And
  Giuseppe Marra \\
  KU Leuven \\
  \texttt{giuseppe.marra@kuleuven.be} \\
  \And
  Francesco Giannini \\
  University of Siena \\
  \texttt{francesco.giannini@unisi.it} \\
  \And
  Michelangelo Diligenti \\
  University of Siena \\
  \texttt{diligmic@diism.unisi.it} \\
  \And
  Zohreh Shams \\
    Babylon Health \\
  University of Cambridge\\
  \texttt{zs315@cam.ac.uk} \\
  \And
  Frederic Precioso \\
  Université Côte d’Azur, Inria, CNRS,\\ I3S, Maasai, Nice, France\\
  \texttt{fprecioso@unice.fr} \\
  \And
  Stefano Melacci \\
  University of Siena\\
  \texttt{mela@diism.unisi.it} \\
  \And
  Adrian Weller \\
  University of Cambridge \\
  Alan Turing Institute \\
  \texttt{aw665@cam.ac.uk} \\
  \And
  Pietro Lio \\
  University of Cambridge \\
  \texttt{pl219@cam.ac.uk} \\
  \And
  Mateja Jamnik \\
  University of Cambridge \\
  \texttt{mateja.jamnik@cl.cam.ac.uk} \\
}

\begin{document}

\maketitle

\begin{abstract}
Deploying AI-powered systems requires trustworthy models supporting effective human interactions, going beyond raw prediction accuracy. Concept bottleneck models promote trustworthiness by conditioning classification tasks on an intermediate level of human-like concepts. This enables human interventions which can correct mispredicted concepts to improve the model's performance. However, existing concept bottleneck models are unable to find optimal compromises between high task accuracy, robust concept-based explanations, and effective interventions on concepts---particularly in real-world conditions where complete and accurate concept supervisions are scarce. To address this, we propose Concept Embedding Models, a novel family of concept bottleneck models which goes beyond the current accuracy-vs-interpretability trade-off by learning interpretable high-dimensional concept representations. Our experiments demonstrate that Concept Embedding Models  (1) attain better or competitive task accuracy w.r.t. standard neural models without concepts, (2) provide concept representations capturing meaningful semantics including and beyond their ground truth labels, (3) support test-time concept interventions whose effect in test accuracy surpasses that in standard concept bottleneck models, and (4) scale to real-world conditions where complete concept supervisions are scarce.
\end{abstract}

\section{Introduction}
What is an \textit{apple}? While any child can explain what an ``apple'' is by enumerating its characteristics, deep neural networks (DNNs) fail to explain what they learn in human-understandable terms despite their high prediction accuracy~\citep{arrieta2020explainable}. This accuracy-vs-interpretability trade-off has become a major concern as high-performing DNNs become commonplace in practice~\citep{goddard2017eu,ras2018explanation,rudin2021interpretable}, thus questioning the ethical~\citep{duran2021afraid, lo2020ethical} and legal~\citep{wachter2017counterfactual, gdpr2017} ramifications of their deployment.

\begin{figure}[!h]
    \centering
    \begin{subfigure}[b]{0.32\textwidth}
        \includegraphics[width=\textwidth]{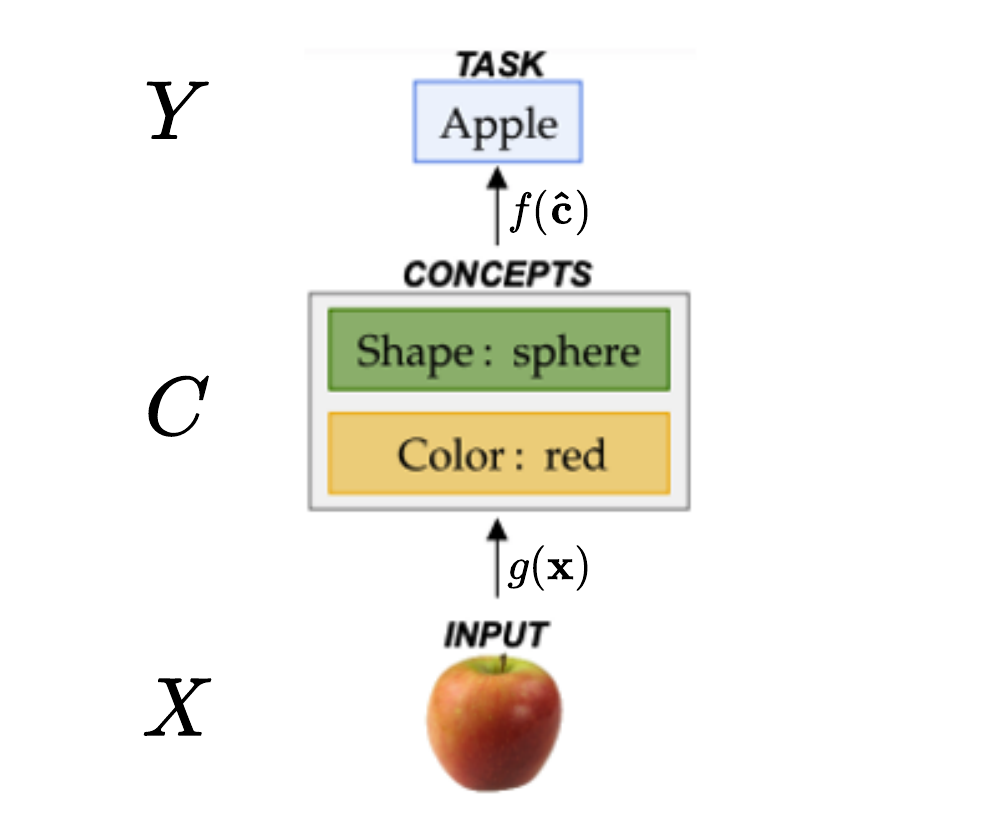}
        \subcaption{}
        \label{fig:pre_results_cbm}
    \end{subfigure}
    \begin{subfigure}[b]{0.32\textwidth}
        \includegraphics[width=\textwidth]{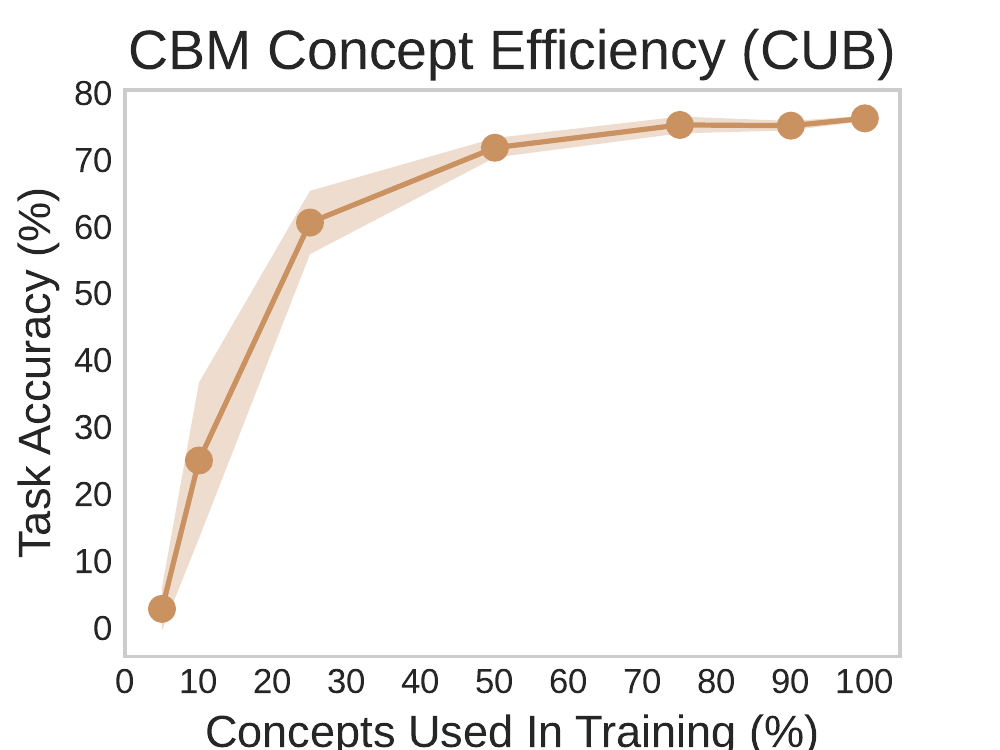}
        \subcaption{}
        \label{fig:pre_results_efficiency}
    \end{subfigure}
    \begin{subfigure}[b]{0.32\textwidth}
        \includegraphics[width=0.93\textwidth]{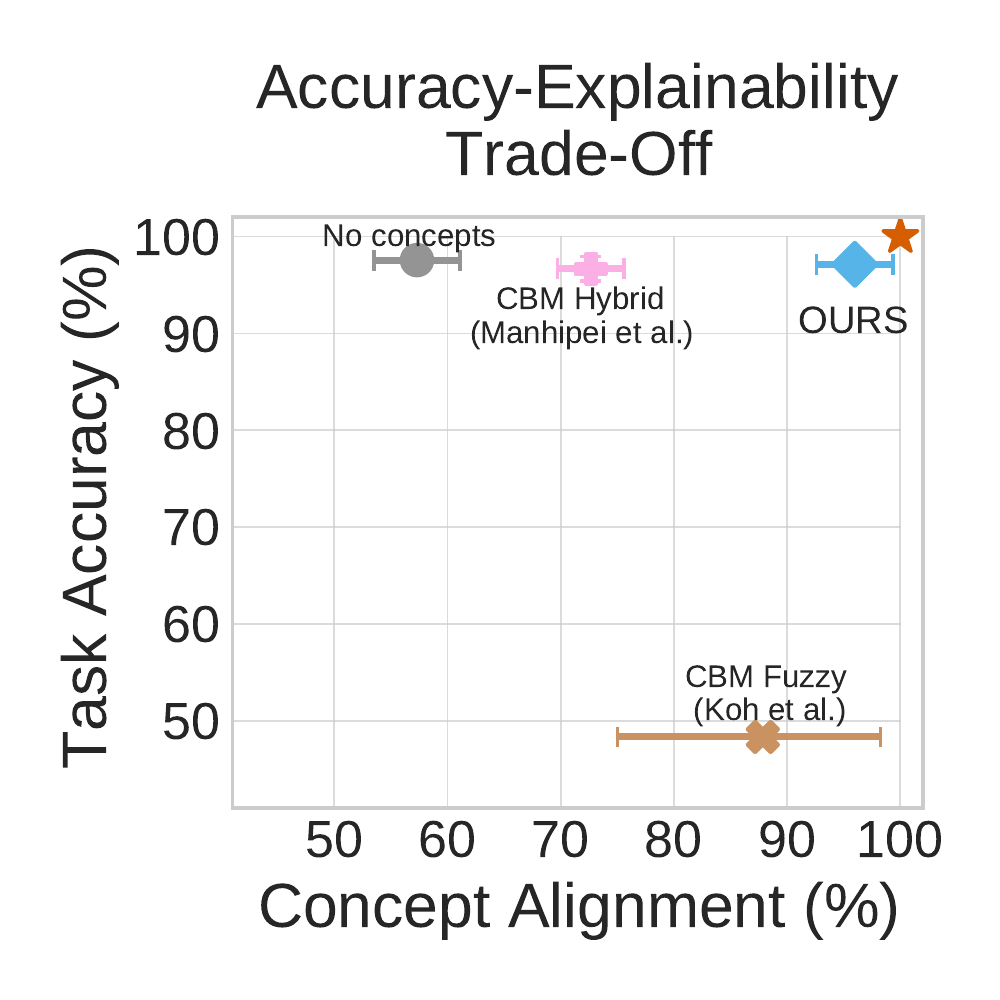}\vspace*{-3mm}
        \subcaption{}
        \label{fig:pre_results_tradeoff}
    \end{subfigure}
    \caption{(a) A concept bottleneck model, (b) task accuracy after using only a fraction of total concept annotations to train a CBM on CUB~\citep{wah2011caltech} and, (c) the accuracy-vs-interpretability trade-off (corner star represents the optimal trade-off).}
    \label{fig:pre_results}
\end{figure}

Concept bottleneck models (CBMs,~\citep{koh2020concept}, Figure~\ref{fig:pre_results_cbm}) aim at replacing ``black-box'' DNNs by first learning to predict a set of concepts, that is, ``interpretable'' high-level units of information (e.g., ``color'' or ``shape'')~\citep{ghorbani2019interpretation} provided at training time~\citep{kim2018interpretability, chen2020concept}, and then using these concepts to learn a downstream classification task. Predicting tasks as a function of concepts engenders user trust~\citep{shen2022trust} by allowing predictions to be explained in terms of concepts and by supporting human interventions, where at test-time an expert can correct a mispredicted concept, possibly changing the CBM's output. That said, concept bottlenecks may impair task accuracy~\citep{koh2020concept, mahinpei2021promises}, especially when concept labels do not contain all the necessary information for accurately predicting a downstream task (i.e., they form an ``incomplete'' representation of the task~\citep{yeh2020completeness}), as seen in Figure~\ref{fig:pre_results_efficiency}. In principle, extending a CBM's bottleneck with a set of unsupervised neurons may improve task accuracy, as observed by~\citet{mahinpei2021promises}. However, as we will demonstrate in this work, such a hybrid approach not only significantly hinders the performance of concept interventions, but it also affects the interpretability of the learnt bottleneck, thus undermining user trust~\citep{shen2022trust}. Therefore, we argue that novel concept-based architectures are required to overcome the current accuracy/interpretability pitfalls of CBMs, thus enabling their deployment in real-world settings where concept annotations are likely to be incomplete.

In this paper, we propose Concept Embedding Models (CEMs), a novel concept bottleneck model (described in Section~\ref{sec:nesy_emb}) which overcomes the current accuracy-vs-interpretability trade-off found in concept-incomplete settings (as shown in Figure~\ref{fig:pre_results_tradeoff}). Furthermore, we introduce two new metrics for evaluating concept representations (Section~\ref{sec:metrics}) and use them to help understand why our approach circumvents the limits found in the current state-of-the-art CBMs (Section~\ref{sec:experiments}).  Our experiments show that CEM (1) attains better or competitive task accuracy w.r.t. standard DNNs trained without concept supervision, (2) learns concept representations that capture meaningful semantics at least as well as vanilla CBMs, and (3) supports effective test-time concept interventions.

\section{Background}
\label{sec:background}

\paragraph{Concept bottleneck models (CBMs, \cite{koh2020concept})} A concept bottleneck model learns a mapping from samples $\mathbf{x} \in X$ to labels $y \in Y$ by means of: (i) a concept encoder function $g: X \rightarrow C$ which maps samples from the input space $\mathbf{x} \in X \subseteq \mathbb{R}^n$ to an intermediate space $\mathbf{\hat{c}} \in C \subseteq \mathbb{R}^{k}$ formed by $k$ concepts, and (ii) a label predictor function $f: C \rightarrow Y$ which maps samples from the concept space $\mathbf{\hat{c}} \in C$ to a downstream task space $\mathbf{\hat{y}} \in Y \subseteq \mathbb{R}^l$. A CBM requires a dataset composed of tuples in $X \times \mathcal{C} \times \mathcal{Y}$, where each sample consists of input features $\mathbf{x}$ (e.g., an image's pixels),  ground truth concept vector $\mathbf{c} \in \{0, 1\}^k$ (i.e., a binary vector where each entry represents whether a concept is active or not) and a task label $y$ (e.g., an image's class). During training, a CBM is encouraged to align $\mathbf{\hat{c}} = g(\mathbf{x})$ and $\mathbf{\hat{y}} = f(g(\mathbf{x}))$ to $\mathbf{x}$'s corresponding ground truth concepts $\mathbf{c}$ and task labels $y$, respectively. This can be done by (i) \textit{sequentially} training first the concept encoder and then using its output to train the label predictor, (ii) \textit{independently} training the concept encoder and label predictor and then combining them to form a CBM, or (iii) \textit{jointly} training the concept encoder and label predictor via a weighted sum of cross entropy losses.

\paragraph{Concept representations in CBMs} For each sample $\mathbf{x} \in X$, the concept encoder $g$ learns $k$ different scalar concept representations $\hat{c}_1,\ldots,\hat{c}_k$. Boolean and Fuzzy CBMs~\citep{koh2020concept} assume that each dimension of $\mathbf{\hat{c}}$, which we describe by $\hat{c}_i = s(\hat{\textbf{c}})_{[i]} \in [0,1]$, is aligned with a single ground truth concept and represents a probability of that concept being active. The element-wise activation function $s: \mathbb{R} \rightarrow [0, 1]$ can be either a thresholding function $s(x) \triangleq \mathbb{1}_{x \ge 0.5}$ (\textit{Boolean} CBM) or sigmoidal function $s(x) \triangleq 1/(1 + e^{-x})$ (\textit{Fuzzy} CBM).\footnote{In practice (e.g., \citep{koh2020concept}) one may use logits rather than sigmoidal activations to improve gradient flow~\citep{glorot2011deep}.} A natural extension of this framework is a \textit{Hybrid} CBM~\citep{mahinpei2021promises}, where $\mathbf{\hat{c}} \in \mathbb{R}^{(k + \gamma)}$ contains $\gamma$ unsupervised dimensions and $k$ supervised concept dimensions which, when concatenated, form a shared concept vector (i.e., an ``embedding'').

\paragraph{Concept Interventions in CBMs} 
Interventions are one of the core motivations behind CBMs~\cite{koh2020concept}. Through interventions, concept bottleneck models allow experts to improve a CBM's task performance by rectifying mispredicted concepts by setting, at test-time, $\hat{c}_i := c_i$ (where $c_i$ is the ground truth value of the $i$-th concept).
Such interventions can significantly improve CBMs performance within a human-in-the-loop setting~\cite{koh2020concept}. Furthermore, interventions enable the construction of meaningful concept-based counterfactuals~\citep{wachter2017counterfactual}. For example, intervening on a CBM trained to predict bird types from images can determine that when the size of a ``black'' bird with ``black'' beak changes from ``medium'' to ``large'', while all other concepts remain constant, then one may classify the bird as a ``raven'' rather than a ``crow''.

\section{Concept Embedding Models} \label{sec:nesy_emb}

In real-world settings, where complete concept annotations are costly and rare, vanilla CBMs may need to compromise their task performance in order to preserve their interpretability~\citep{koh2020concept}.
While Hybrid CBMs are able to overcome this issue by adding extra capacity in their bottlenecks, this comes at the cost of their interpretability and their responsiveness to concept interventions, thus undermining user trust~\citep{shen2022trust}. To overcome these pitfalls, we propose \textit{Concept Embedding Models} (CEMs), a concept-based architecture which represents each concept as a supervised vector. Intuitively, using high-dimensional embeddings to represent each concept allows for extra \textit{supervised} learning capacity, as opposed to Hybrid models where the information flowing through their \textit{unsupervised} bottleneck activations is concept-agnostic. In the following section, we introduce our architecture and describe how it learns  a mixture of two semantic embeddings for each concept (Figure~\ref{fig:split_emb_architecture}). We then discuss how interventions are performed in CEMs and introduce \textit{RandInt}, a train-time regularisation mechanism that incentivises our model to positively react to interventions at test-time.

\subsection{Architecture}
\label{sec:model}
For each concept, CEM learns a mixture of two embeddings with explicit semantics representing the concept's activity. Such design allows our model to construct evidence both in favour of and against a concept being active, and supports simple concept interventions as one can switch between the two embedding states at intervention time.

\begin{figure}[t]
    \centering
    \includegraphics[width=0.95\textwidth]{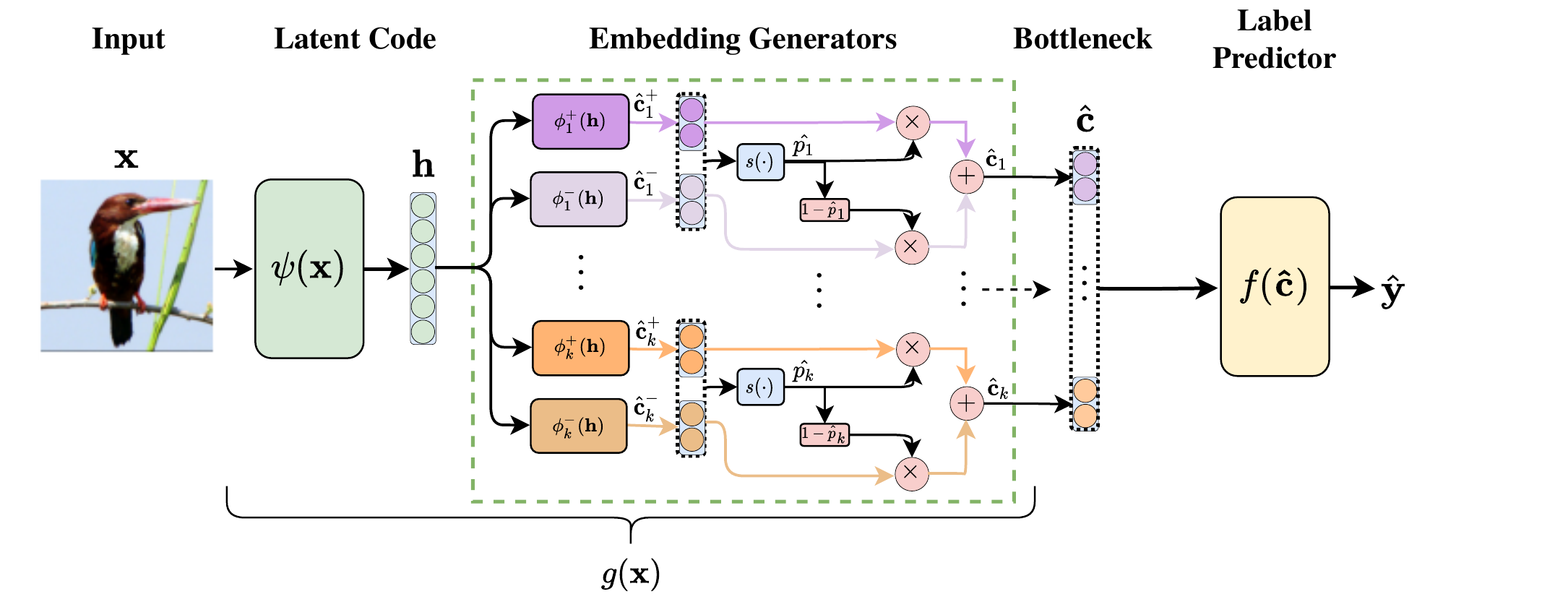}
    \caption{\textbf{Concept Embedding Model}: from an intermediate latent code $\mathbf{h}$, we learn two embeddings per concept, one for when it is active (i.e., $\hat{\textbf{c}}^+_i$), and another when it is inactive (i.e., $\hat{\textbf{c}}^-_i$). Each concept embedding (shown in this example as a vector with $m=2$ activations) is then aligned to its corresponding ground truth concept through the scoring function $s(\cdot)$, which learns to assign activation probabilities $\hat{p}_i$ for each concept. These probabilities are used to output an embedding for each concept via a weighted mixture of each concept's positive and negative embedding.
    }
    \label{fig:split_emb_architecture}
\end{figure}

We represent concept $c_i$ with two embeddings $\hat{\textbf{c}}^+_i, \hat{\textbf{c}}^-_i \in \mathbb{R}^m$, each with a specific semantics: $\hat{\textbf{c}}^+_i$ represents its active state (concept is \texttt{true}) while $\hat{\textbf{c}}^-_i$ represents its inactive state (concept is \texttt{false}). To this aim, a DNN $\psi(\mathbf{x})$ learns a latent representation $\mathbf{h} \in \mathbb{R}^{n_\text{hidden}}$ which is the input to CEM's embedding generators. CEM then feeds $\mathbf{h}$ into two concept-specific fully connected layers, which learn two concept embeddings in $\mathbb{R}^m$, namely $\hat{\mathbf{c}}^+_i = \phi^+_i(\mathbf{h}) = a(W^+_i\mathbf{h} + \mathbf{b}^+_i$) and $\hat{\mathbf{c}}^-_i = \phi^-_i(\mathbf{h}) = a(W^-_i\mathbf{h} + \mathbf{b}^-_i)$.\footnote{In practice, we use a leaky-ReLU for the activation $a(\cdot)$.} Notice that while more complicated models can be used to parameterise our concept embedding generators $\phi^+_i(\mathbf{h})$ and $\phi^-_i(\mathbf{h})$, we opted for a simple one-layer neural network to constrain parameter growth in models with large bottlenecks.
Our architecture encourages embeddings $\hat{\mathbf{c}}^+_i$ and $\hat{\mathbf{c}}^-_i$ to be aligned with ground-truth concept $c_i$ via a learnable and differentiable scoring function $s: \mathbb{R}^{2 m} \rightarrow [0, 1]$, trained to predict the probability $\hat{p}_i \triangleq s([\hat{\mathbf{c}}^+_i, \hat{\mathbf{c}}^-_i]^T) =  \sigma\big(W_s[\hat{\mathbf{c}}^+_i, \hat{\mathbf{c}}^-_i]^T + \mathbf{b}_s\big)$ of concept $c_i$ being active from the embeddings' joint space. For the sake of parameter efficiency, parameters $W_s$ and $\mathbf{b}_s$ are shared across all concepts. Once both semantic embeddings are computed, we construct the final concept embedding $\hat{\mathbf{c}}_i$ for $c_i$ as a weighted mixture of $\hat{\mathbf{c}}^+_i$ and $\hat{\mathbf{c}}^-_i$:
\[
\hat{\mathbf{c}}_i \triangleq \big(\hat{p}_i \hat{\mathbf{c}}^+_i + (1 - \hat{p}_i) \hat{\mathbf{c}}^-_i \big)
\]
Intuitively, this serves a two-fold purpose: (i) it forces the model to depend only on $\hat{\mathbf{c}}^+_i$ when the $i$-th concept is active, that is, $c_i = 1$ (and only on $\hat{\mathbf{c}}^-_i$ when inactive), leading to two different semantically meaningful latent spaces, and (ii) it enables a clear intervention strategy where one switches the embedding states when correcting a mispredicted concept, as discussed below.
Finally, all $k$ mixed concept embeddings are concatenated, resulting in a bottleneck $g(\mathbf{x}) = \hat{\textbf{c}}$ with $k\cdot m$ units (see end of Figure~\ref{fig:split_emb_architecture}). This is passed to the label predictor $f$ to obtain a downstream task label. In practice, following~\citet{koh2020concept}, we use an interpretable label predictor $f$ parameterised by a simple linear layer, though more complex functions could be explored too. Notice that as in vanilla CBMs, CEM provides a concept-based explanation for the output of $f$ through its concept probability vector $\hat{\mathbf{p}}(\mathbf{x}) \triangleq [\hat{p}_1, \cdots, \hat{p}_k ]$, indicating the predicted concept activity. This architecture can be trained in an end-to-end fashion by \textit{jointly} minimising via stochastic gradient descent a weighted sum of the cross entropy loss on both task prediction and concept predictions:
\begin{align}
    \mathcal{L} \triangleq \mathbb{E}_{(\mathbf{x}, y, \mathbf{c})}\Big[ \mathcal{L}_\text{task}\Big(y, f\big(g(\mathbf{x})\big)\Big) + \alpha \mathcal{L}_\text{CrossEntr}\Big(\mathbf{c}, \hat{\mathbf{p}}(\mathbf{x})\Big) \Big]
\end{align}
where hyperparameter $\alpha \in \mathbb{R}^+$ controls the relative importance of concept and task accuracy.

\subsection{Intervening with Concept Embeddings}
\label{sec:randint}
As in vanilla CBMs, CEMs support test-time concept interventions. To intervene on concept $c_i$, one can update $\hat{\mathbf{c}}_i$ by swapping the output concept embedding for the one semantically aligned with the concept ground truth label. For instance, if for some sample $\mathbf{x}$ and concept $c_i$ a CEM predicted $\hat{p}_i = 0.1$ while a human expert knows that concept $c_i$ is active ($c_i=1$), they can perform the intervention $\hat{p}_i := 1$. This operation updates CEM's bottleneck by setting $\hat{\mathbf{c}}_i$ to $\hat{\mathbf{c}}^+_i$ rather than $\big(0.1 \hat{\mathbf{c}}^+_i + 0.9 \hat{\mathbf{c}}^-_i\big)$. Such an update allows the downstream label predictor to act on information related to the corrected concept.
In addition, we introduce \textit{RandInt}, a regularisation strategy exposing CEMs to concept interventions during training to improve the effectiveness of such actions at test-time. RandInt randomly performs independent concept interventions during training with probability $p_\text{int}$ (i.e., $\hat{p}_i$ is set to $\hat{p}_i := c_i$ for concept $c_i$ with probability $p_\text{int}$). In other words, for all concepts $c_i$, during training we compute embedding $\hat{\mathbf{c}}_i$ as:
\[
    \hat{\mathbf{c}}_i = \begin{cases}
        \big(c_i \hat{\mathbf{c}}^+_i + (1 - c_i) \hat{\mathbf{c}}^-_i\big) & \text{with probability } p_\text{int} \\
        \big(\hat{p}_i \hat{\mathbf{c}}^+_i + (1 - \hat{p}_i) \hat{\mathbf{c}}^-_i\big) & \text{with probability } (1 - p_\text{int})
    \end{cases}
\]

while at test-time we always use the predicted probabilities for performing the mixing. During backpropagation, this strategy forces feedback from the downstream task to update only the correct concept embedding (e.g., $\hat{\mathbf{c}}^+_i$ if $c_i = 1$) while feedback from concept predictions updates both $\hat{\mathbf{c}}^+_i$ and $\hat{\mathbf{c}}^-_i$. Under this view, RandInt can be thought of as learning an average over an exponentially large family of CEM models (similarly to dropout~\citep{srivastava2014dropout}) where some of the concept representations are trained using only feedback from their concept label while others receive training feedback from both their concept and task labels.

\section{Evaluating concept bottlenecks} \label{sec:metrics}

To the best of our knowledge, while a great deal of attention has been paid to concept-based explainability in recent years, existing work still fails to provide methods that can be used to evaluate the interpretability of a concept embedding or to explain why certain CBMs underperform in their task predictions. With this in mind, we propose (i) a new metric for evaluating concept quality in multidimensional representations and (ii) an information-theoretic method which, by analysing the information flow in concept bottlenecks, can help understand why a CBM may underperform in a downstream task.

\paragraph{Concept Alignment Score (CAS)} 
The Concept Alignment Score (CAS) aims to measure how much learnt concept representations can be trusted as faithful representations of their ground truth concept labels. Intuitively, CAS generalises concept accuracy by considering the homogeneity of predicted concept labels within groups of samples which are close in a concept subspace. More specifically, for each concept $c_i$ the CAS applies a clustering algorithm $\kappa$ to find $\rho > 2$ clusters, assigning to each sample $\mathbf{x}^{(j)}$ a cluster label $\pi_i^{(j)} \in \{1, \cdots, \rho\}$. We compute this label by clustering samples using their $i$-th concept representations $\{\hat{\textbf{c}}_i^{(1)}, \hat{\textbf{c}}_i^{(2)}, \cdots\}$. Given $N$ test samples, the homogeneity score $h(\cdot)$~\citep{rosenberg2007v} then computes the conditional entropy $H$ of ground truth labels $C_i = \{c_i^{(j)}\}_{j=1}^{N}$ w.r.t. cluster labels $\Pi_i(\kappa, \rho) = \{\pi_i^{(j)}\}_{j=1}^{N}$, that is, $h = 1$ when $H(C_i,\Pi_i)=0$ and $h = 1 - H(C_i, \Pi_i)/H(C_i)$ otherwise. The higher the homogeneity, the more a learnt concept representation is ``aligned'' with its labels, and can thus be trusted as a faithful representation. CAS averages homogeneity scores over all concepts and number of clusters $\rho$, providing a normalised score in $[0,1]$:
\begin{equation}
\label{eq:cas}
    \text{CAS}(\mathbf{\hat{c}}_1, \cdots, \mathbf{\hat{c}}_k) \triangleq \frac{1}{N - 2}\sum_{\rho=2}^N \Bigg(\frac{1}{k} \sum_{i=1}^k h(C_i, \Pi_i(\kappa, \rho)) \Bigg)
\end{equation}
To tractably compute CAS in practice, we sum homogeneity scores by varying $\rho$ across $\rho \in \{2, 2 + \delta, 2 + 2 \delta, \cdots, N\}$ for some $\delta > 1$ (details in Appendix~\ref{sec:appendix_cas}). Furthermore, we use k-Medoids~\citep{kaufman1990partitioning} for cluster discovery, as used in~\citet{ghorbani2019interpretation} and~\citet{magister2021gcexplainer}, and use concept logits when computing the CAS for Boolean and Fuzzy CBMs. For Hybrid CBMs, we use $\hat{\mathbf{c}}_i \triangleq [\hat{\mathbf{c}}_{[k:k + \gamma]}, \hat{\mathbf{c}}_{[i:(i + 1)]}]^T$ as the concept representation for $c_i$ given that the extra capacity is a shared embedding across all concepts.

\paragraph{Information bottleneck} 
The relationship between the quality of concept representations w.r.t. the input distribution remains widely unexplored. Here we propose to analyse this relationship using information theory methods for DNNs developed in~\citet{tishby2000information} and~\citet{tishby2015deep}. In particular, we compare concept bottlenecks using the Information Plane method~\citep{tishby2000information} to study the information flow at concept level. To this end, we measure the evolution of the Mutual Information ($I(\cdot, \cdot)$) of concept representations w.r.t. the input and output distributions across training epochs. We conjecture that embedding-based CBMs circumvent the information bottleneck by preserving more information than vanilla CBMs from the input distribution as part of their high-dimensional activations. If true, such effect should be captured by Information Planes in the form of a positively correlated evolution of $I(X, \hat{C})$, the Mutual Information (MI) between inputs $X$ and learnt concept representations $\hat{C}$, and $I(\hat{C}, Y)$, the MI between learnt concept representations $\hat{C}$ and task labels $Y$. In contrast, we anticipate that scalar-based concept representations (e.g., Fuzzy and Bool CBMs), will be forced to compress the information from the input data at concept level, leading to a compromise between the $I(X, \hat{C})$ and $I(\hat{C}, Y)$. Further details on our implementation are in Appendix~\ref{sec:appendix_mi}.

\section{Experiments}
\label{sec:experiments}

In this section, we address the following research questions:
\begin{itemize}
    \item \textbf{Task accuracy ---} What is the impact of concept embeddings on a CBM's downstream task performance? Are models based on concept embeddings still subject to an information bottleneck~\citep{tishby2000information}?
    \item \textbf{Interpretability ---} Are CEM concept-based explanations aligned with ground truth concepts? Do they offer interpretability beyond simple concept prediction and alignment?
    \item \textbf{Interventions ---} Do CEMs allow meaningful concept interventions when compared to Hybrid or vanilla CBMs?
\end{itemize}

\subsection{Setup}
\label{sec:setup}

\paragraph{Datasets}
For our evaluation, we propose three simple benchmark datasets of increasing concept complexity (from Boolean to vector-based concepts): (1) \textit{XOR} (inspired by~\cite{minsky2017perceptrons}) in which tuples $(x_1, x_2) \in [0, 1]^2$ are annotated with two Boolean concepts $\{ \mathbb{1}_{c_i > 0.5}\}_{i=1}^2$ and labeled as $y = c_1 \text{ XOR } c_2$; (2) \textit{Trigonometric} (inspired by~\cite{mahinpei2021promises}) in which three latent normal random variables $\{b_i\}_{i=1}^3$ are used to generate a 7-dimensional input whose three concept annotations are a Boolean function of $\{b_i\}_{i=1}^3$ and task label is a linear function of the same; (3) \textit{Dot} in which four latent random vectors  $\mathbf{v}_1, \mathbf{v}_2, \mathbf{w}_1, \mathbf{w}_2 \in \mathbb{R}^2$ are used to generate two concept annotations, representing whether latent vectors $\mathbf{v}_i$ point in the same direction of reference vectors $\mathbf{w}_i$, and task labels, representing whether the two latent vectors $\mathbf{v}_1$ and $\mathbf{v}_2$ point in the same direction. Furthermore, we evaluate our methods on two real-world image tasks: the Caltech-UCSD Birds-200-2011 dataset (CUB,~\citep{wah2011caltech}), preprocessed as in~\citep{koh2020concept}, and the Large-scale CelebFaces Attributes dataset (CelebA,~\citep{liu2015deep}). In our CUB task we have $112$ complete concept annotations and $200$ task labels while in our CelebA task we construct $6$ balanced incomplete concept annotations and each image can be one of $256$ classes. Therefore, we use CUB to test each model in a real-world task where concept annotations are numerous and they form a complete description of their downstream task. In contrast, our CelebA task is used to evaluate the behaviour of each method in scenarios where the concept annotations are scarce and incomplete w.r.t. their downstream task. Further details on these datasets and their properties are provided in Appendix~\ref{sec:appendix_data_details}.

\paragraph{Baselines} We compare CEMs against Bool, Fuzzy, and Hybrid Joint-CBMs as they all provide concept-based explanations for their predictions and allow concept interventions at test-time. Note that this set excludes architectures such as Self-Explainable Neural Networks~\citep{alvarez2018towards} and Concept Whitening~\cite{chen2020concept} as they do not offer a clear mechanism for intervening on their concept bottlenecks. To ensure fair comparison, we use the same architecture capacity across all models
\reb{. We empirically justify this decision in Appendix~\ref{sec:appendix_resnet18} by showing that the results discussed in this section do not change if one modifies the underlying model architecture.}
Similarly, we use the same values of $\alpha$ and $m$ within a dataset for all models trained on that dataset and set $p_\text{int} = 0.25$ when using CEM (see Appendix~\ref{sec:appendix_prob_ablation} for an ablation study of $p_\text{int}$ showing how intervention improvement plateaus around this value). When using Hybrid CBMs, we include as many activations in their bottlenecks as their CEM counterparts \reb{(so that they both end up with a bottleneck with $k m $ activations)} and use a Leaky-ReLU activation for unsupervised activations.
Finally, in our evaluation we include a DNN without concept supervision with the same capacity as its CEM counterpart to measure the effect of concept supervision in our model's performance. 
For further details on the model architectures and training hyperparameters, please refer to Appendix~\ref{sec:appendix_architectures}.

\paragraph{Metrics}
We measure a model's performance based on four metrics. First, we measure task and concept classification performance in terms of both \textit{task and mean concept accuracy}. Second, we evaluate the interpretability of learnt concept representations via our \textit{concept alignment score}. To easily visualise the accuracy-vs-interpretability trade-off, we plot our results in a two-dimensional plane showing both task accuracy and concept alignment. Third, we study the information bottleneck in our models via \textit{mutual information} (MI) and the Information Plane technique~\citep{shwartz2017opening}. Finally, we quantify user trust~\citep{shen2022trust} by evaluating a model's task performance after concept interventions.  All metrics in our evaluation, across all experiments, are computed on test sets using $5$ random seeds, from which we compute a metric's mean and $95\%$ confidence interval using the Box-Cox transformation for non-normal distributions. 

\subsection{Task Accuracy}

\begin{figure}[t]
    \centering
    \includegraphics[width=0.95\textwidth]{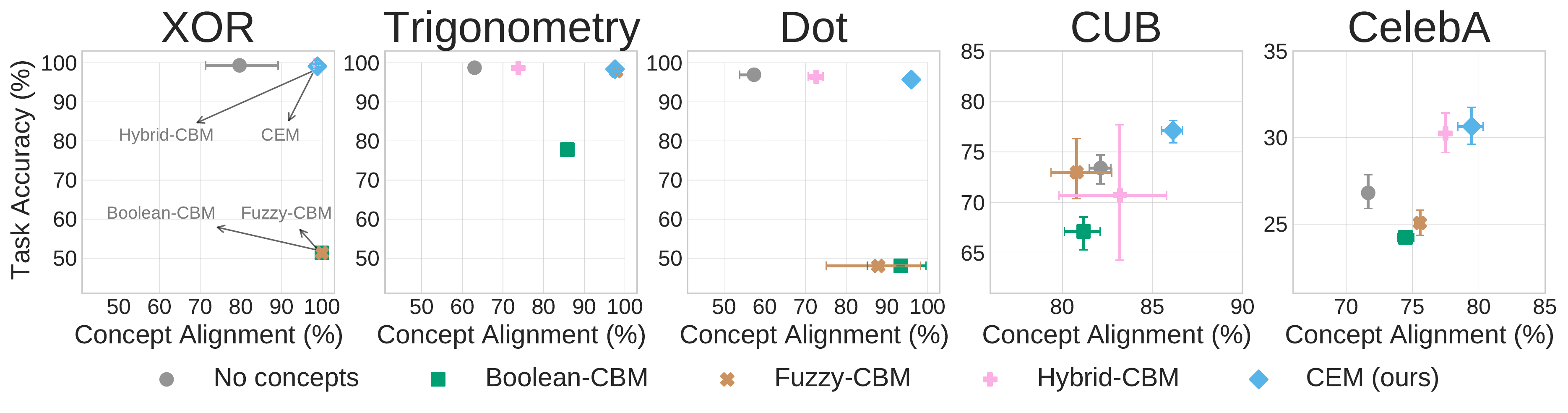}
    \caption{Accuracy-vs-interpretability trade-off in terms of \textbf{task accuracy} and \textbf{concept alignment score} for different concept bottleneck models. In CelebA, our most constrained task, we show the top-1 accuracy for consistency with other datasets.}
    \label{fig:accuracy}
\end{figure}

\paragraph{CEM improves generalisation accuracy (y-axis of Figure \ref{fig:accuracy})}
Our evaluation shows that embedding-based CBMs (i.e., Hybrid-CBM and CEM) can achieve competitive or better downstream accuracy than DNNs that do not provide any form of concept-based explanations, and can easily outperform Boolean and Fuzzy CBMs by a large margin (up to $+45\%$ on Dot). This effect is emphasised when the downstream task is not a linear function of the concepts (e.g., XOR and Trigonometry) or when concept annotations are incomplete (e.g., Dot and CelebA). At the same time, we observe that all concept-based models achieve a similar high mean concept accuracy across all datasets (see Appendix~\ref{sec:appendix_taks_and_concept_perf}). This suggests that, as hypothesised, the trade-off between concept accuracy and task performance in concept-incomplete tasks is significantly alleviated by the introduction of concept embeddings in a CBM's bottleneck. \reb{Similar results can be observed when training our baselines using only a fraction of the available concepts in CUB as seen in Appendix~\ref{sec:appendix_concept_subsampling}.}
Finally, notice that CelebA showcases how including concept supervision during training (as in CEM) can lead to an even higher task accuracy than the one obtained by a vanilla end-to-end model ($+5\%$ compared to ``No concepts'' model). This result further suggests that concept embedding representations enable high levels of interpretability without sacrificing performance.

\begin{figure}[!ht]
    \centering
    \includegraphics[width=0.95\textwidth]{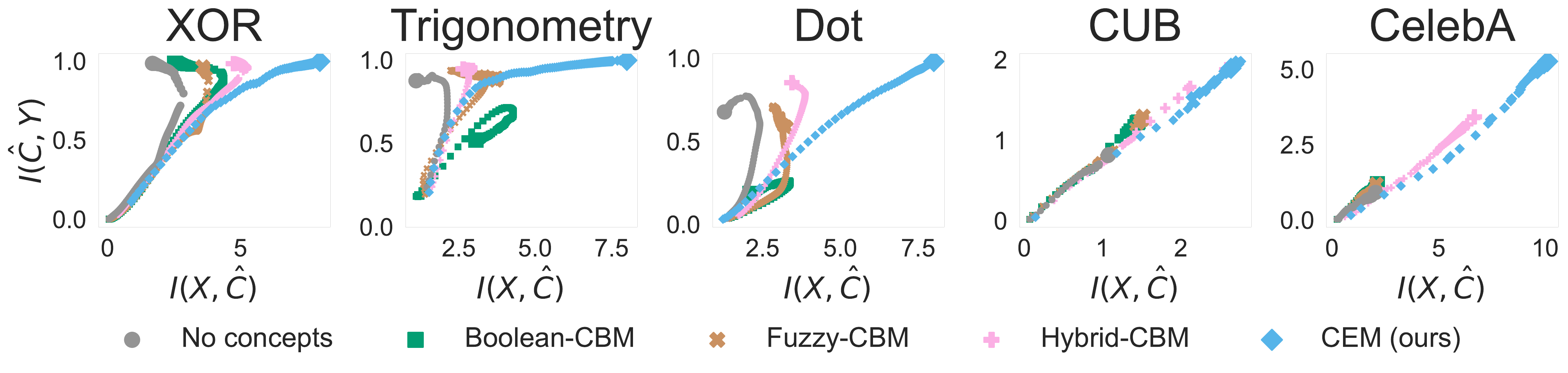}
    \caption{Mutual Information (MI) of concept representations ($\hat{C}$) w.r.t. input distribution ($X$) and ground truth labels ($Y$) during training.
    The size of the points is proportional to the training epoch.
    }
    \label{fig:info_plane}
\end{figure}

\paragraph{CEM overcomes the information bottleneck (Figure~\ref{fig:info_plane})}
The Information Plane method indicates, as hypothesised, that embedding-based CBMs (i.e., Hybrid-CBM and CEM) do not compress input data information, with $I(X, \hat{C})$ monotonically increasing during training epochs. On the other hand, Boolean and Fuzzy CBMs, as well as vanilla end-to-end models, tend to ``forget''~\citep{shwartz2017opening} input data information in their attempt to balance competing objective functions. Such a result constitutes a plausible explanation as to why embedding-based representations are able to maintain both high task accuracy and mean concept accuracy compared to CBMs with scalar concept representations. In fact, the extra capacity allows CBMs to maximise concept accuracy without over-constraining concept representations, thus allowing useful input information to pass by. In CEMs all input information flows through concepts, as they supervise the whole concept embedding. In contrast with Hybrid models, this makes the downstream tasks completely dependent on concepts, which explains the higher concept alignment scores obtained by CEM (as discussed in the next subsection).

\subsection{Interpretability}

\paragraph{CEM learns more interpretable concept representations (x-axis of Figure~\ref{fig:accuracy})}
Using the proposed CAS metric, we show that concept representations learnt by CEMs have alignment scores competitive or even better (e.g., on CelebA) than the ones of Boolean and Fuzzy CBMs. The alignment score also shows, as hypothesised, that hybrid concept embeddings are the least faithful representations---with alignment scores up to $25\%$ lower than CEM in the Dot dataset. This is due to their unsupervised activations containing information which may not be necessarily relevant to a given concept. This result  is a further evidence for why we expect interventions to be ineffective in Hybrid models (as we show shortly).

\begin{figure}[!t]
    \centering
    \begin{subfigure}[b]{0.3\textwidth}
        \centering
        \includegraphics[width=\textwidth]{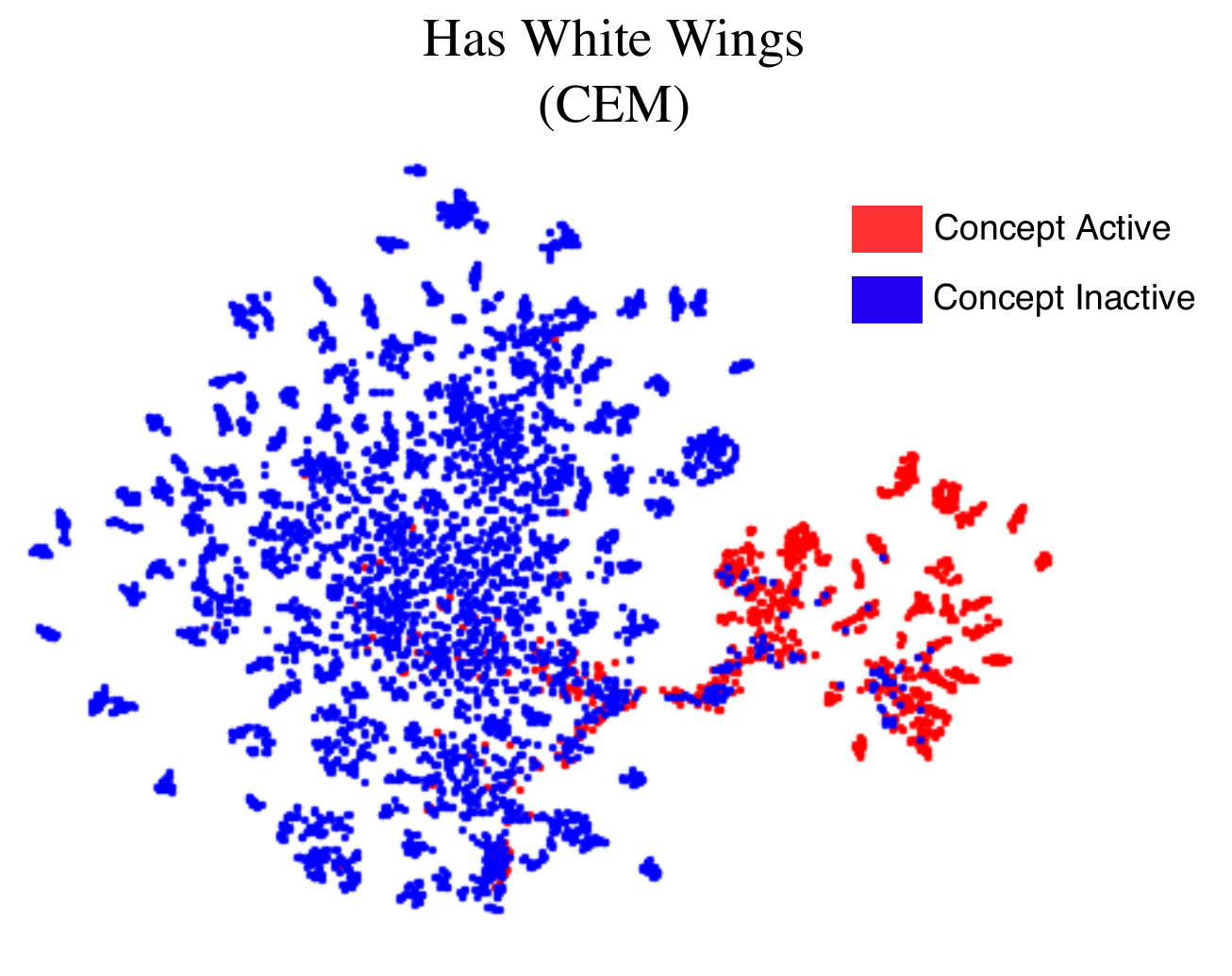}
        \subcaption{
        }
        \label{fig:mixcem_tsne}
    \end{subfigure}
    \begin{subfigure}[b]{0.3\textwidth}
        \centering
        \includegraphics[width=\textwidth]{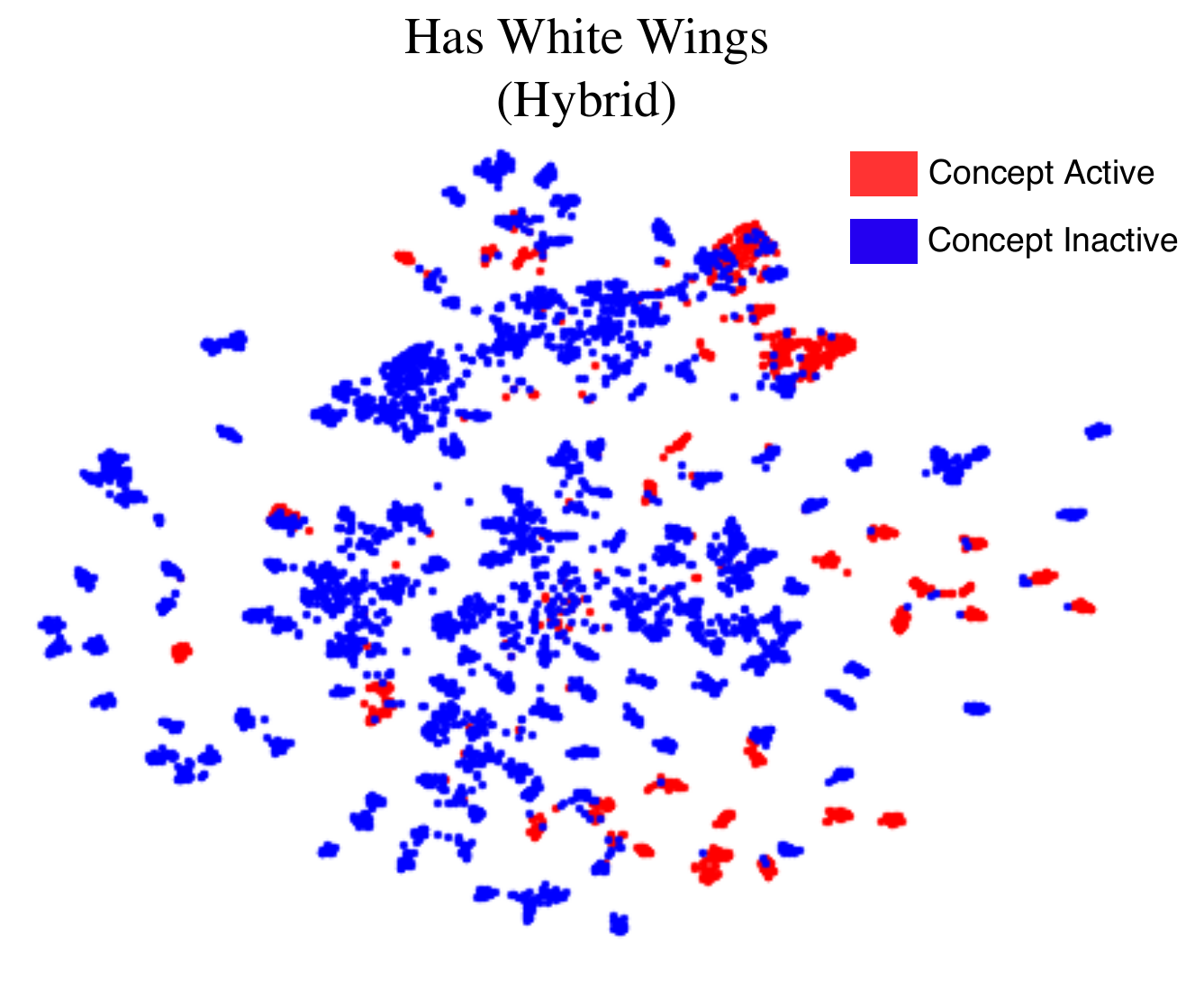}
        \subcaption{
        }
        \label{fig:hybrid_tsne}
    \end{subfigure}
    \begin{subfigure}[b]{0.3\textwidth}
        \centering
        \includegraphics[width=\textwidth]{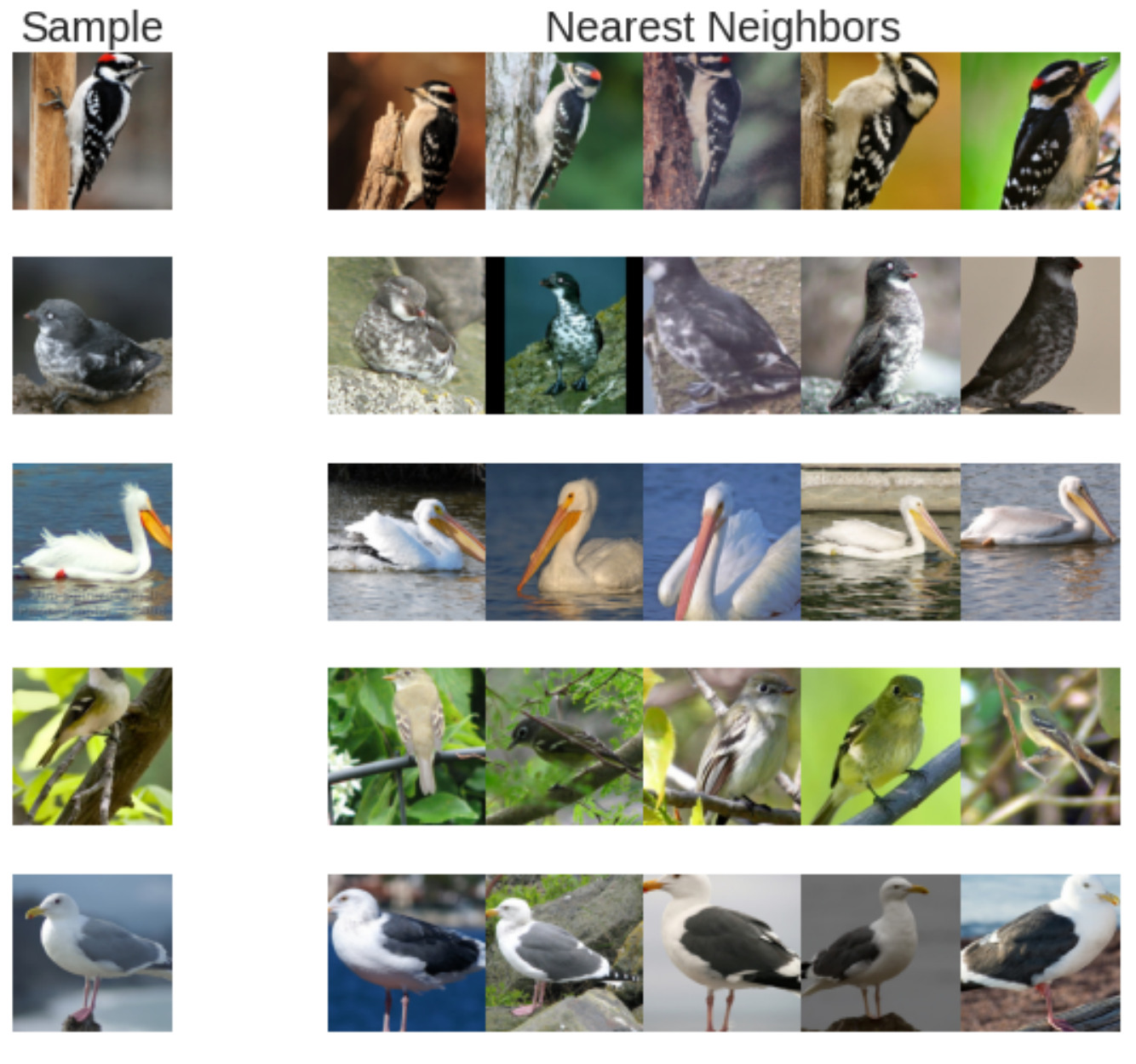}
        \subcaption{
        }
        \label{fig:mixcem_nn}
    \end{subfigure}
    \caption{Qualitative results: (a and b) t-SNE visualisations of ``has white wings'' concept embedding learnt in CUB with sample points coloured red if the concept is active in that sample, (c) top-5 test neighbours of CEM's embedding for the concept ``has white wings'' across 5 random test samples.}
    \label{fig:xai}
\end{figure}

\paragraph{CEM captures meaningful concept semantics (Figure \ref{fig:xai})}
Our concept alignment results hint at the possibility that concept embeddings learnt by CEM may be able to offer more than simple concept prediction. In fact, we hypothesise that their seemingly high alignment may lead to these embeddings forming more interpretable representations than Hybrid embeddings, which can lead to these embeddings serving as better representations for different tasks. To explore this, we train a Hybrid-CBM and a CEM, both with $m=16$, using a variation of CUB with only 25\% of its concept annotations randomly selected before training, resulting in a bottleneck with 28 concepts (see Appendix~\ref{sec:appendix_repr_power_experiment} for details). Once these models have been trained to convergence, we use their learnt bottleneck representations to predict the remaining 75\% of the concept annotations in CUB using a simple logistic linear model. The model trained using the Hybrid bottleneck notably underperfoms when compared to the model trained using the CEM bottleneck (Hybrid-trained model has a mean concept accuracy of 91.83\% $\pm$ 0.51\% while the CEM-trained model's concept accuracy is 94.33\% $\pm$ 0.88\%). This corroborates our CAS results by suggesting that the bottlenecks learnt by CEMs are considerably more interpretable and can therefore serve as powerful feature extractors.

We can further explore this phenomena qualitatively by visualising the embeddings learnt for a single concept using its 2-dimensional t-SNE~\citep{van2008visualizing} plot. As shown in colour in Figure~\ref{fig:mixcem_tsne}, we can see that the embedding space learnt for a concept $\hat{\mathbf{c}}_i$ (here we show the concept ``has white wings'') forms two clear clusters of samples, one for points in which the concept is active and one for points in which the concept is inactive. When performing a similar analysis for the same concept in the Hybrid CBM (Figure~\ref{fig:hybrid_tsne}), where we use the entire extra capacity $\hat{\mathbf{c}}_{[k:k + \gamma]}$ as the concept's embedding representation, we see that this latent space is not as clearly separable as that in CEM's embeddings, suggesting this latent space is unable to capture concept-semantics as clearly as CEM's latent space. Notice that CEM's t-SNE seems to also show smaller subclusters within the activated and inactivated clusters. As Figure~\ref{fig:mixcem_nn} shows, by looking at the nearest Euclidean neighbours in concept's $c_i$ embedding's space, we see that CEM concepts clearly capture a concept's activation, as well as exhibit high class-wise coherence by mapping same-type birds close to each other (explaining the observed subclusters). These results, and similar qualitative results in Appendix~\ref{sec:appendix_qualitative}, suggest \reb{that CEM is learning a hierarchy in its latent space where embeddings are separated with respect to their concept activation and, within the set of embeddings that have the same activation, embeddings are clustered according to their task label.}

\subsection{Interventions}
\label{sec:int_results}
\begin{figure}[!t]
    \centering
    \includegraphics[width=0.9\textwidth]{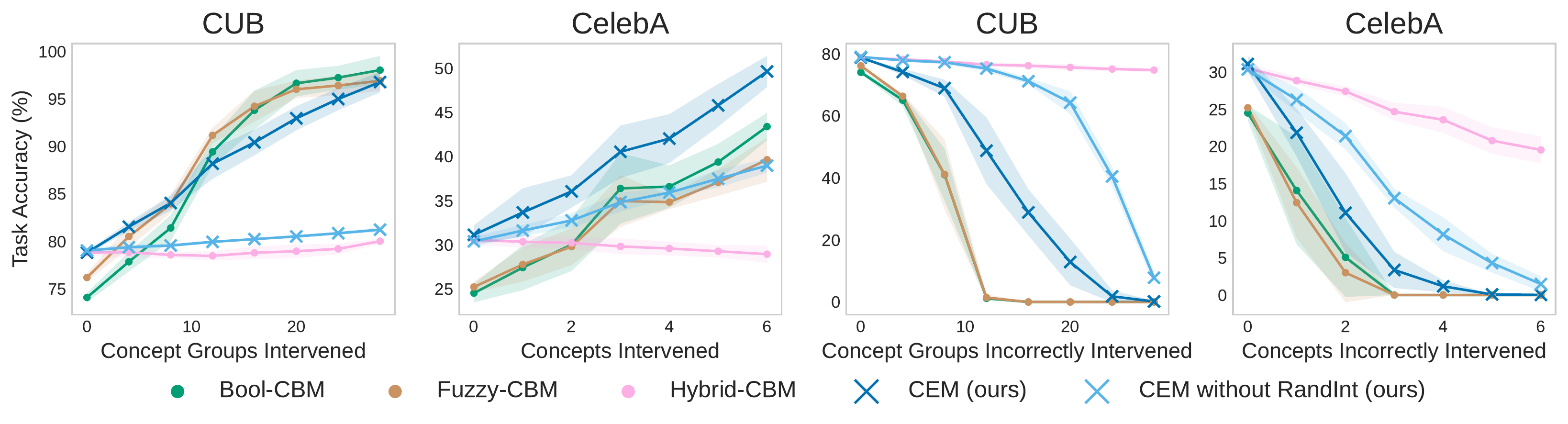}
    \caption{Effects of performing positive random concept interventions (left and center left) and incorrect random interventions (center right and right) for different models in CUB and CelebA. As in~\citep{koh2020concept}, when intervening in CUB we jointly set groups of mutually exclusive concepts.}
    \label{fig:interventions}
\end{figure}

\paragraph{CEM supports effective concept interventions and is more robust to incorrect interventions (Figure~\ref{fig:interventions})} When describing our CEM architecture, we argued in favour of using a mixture of two semantic embeddings for each concept as this would permit test-time interventions which can meaningfully affect entire concept embeddings. In Figure~\ref{fig:interventions} left and center-left, we observe, as hypothesised, that using a mixture of embeddings allows CEMs to be highly responsive to random concept interventions in their bottlenecks. Notice that although all models have a similar concept accuracy (see Appendix~\ref{sec:appendix_taks_and_concept_perf}), we observe that Hybrid CBMs, while highly accurate without interventions, quickly fall short against even scalar-based CBMs once several concepts are intervened in their bottlenecks. In fact, we observe that interventions in Hybrid CBM bottlenecks have little effect on their predictive accuracy, something that did not change if logit concept probabilities were used instead of sigmoidal probabilities. 
\reb{However, even interventions performed by human experts are quite rarely perfect. For this reason, we simulate incorrect interventions (where a concept is set to the wrong value) to measure how robust the model is to such errors. We observe (Figure~\ref{fig:interventions} center-right and right) that CEM's performance deteriorates as more concepts are incorrectly intervened on (as opposed to hybrid-CBMs), while it can withstand a few errors without losing much performance (as opposed to Bool and Fuzzy-CBMs).} 
We suggest that this is a consequence of CEM's ``incorrect'' embeddings still carrying important task-specific information which can then be used by the label predictor to produce more accurate task labels, something worth exploring in future work.
\reb{As a result, users can trust CEMs to better handle a small number of accidental mistakes made by human experts when intervening in its concept activations.}
Finally, by comparing the effect of interventions in both CEMs and CEMs trained without RandInt, we observe that RandInt in fact leads to a model that is not just significantly more receptive to interventions, but is also able to outperform even scalar-based CBMs when large portions of their bottleneck are artificially set by experts (e.g., as in CelebA). \reb{This, as shown in Appendix~\ref{sec:appendix_computational_time}, comes without a significant computational training costs for CEM. 
Interestingly, such a positive effect in concept interventions is not observed if RandInt is used when training our other baselines (see Appendix~\ref{sec:appendix_randint_cbms} for an explanation).}
This suggests that our proposed architecture can not only be trusted in terms of its downstream predictions and concept explanations, as seen above, but it can also be a highly effective model when used along with experts that can correct mistakes in their concept predictions. For further details, including an exploration of performing interventions with Sequential and Independent CBMs, please refer to Appendix~\ref{sec:appendix_intervention}.

\section{Discussion}
\label{sec:discussion}

\paragraph{Relations with the state-of-the-art}
Concept bottleneck models engender user trust~\citep{shen2022trust} by (i) forcing information flow through concept-aligned activations during training and (ii) by supporting human interventions in their concept predictions. This allows CBMs to circumvent the well-known unreliability of post-hoc methods~\citep{adebayo2018sanity, kindermans2019reliability, ghorbani2019interpretation}, such as saliency maps~\citep{sundararajan2017axiomatic, kindermans2017learning, selvaraju2016grad}, and invites their use in settings where input features are naturally hard to reason about (e.g., raw image pixels). In addition, CBMs encourage human interactions allowing experts to improve task performance by rectifying mispredicted concepts, which contrasts other concept-based interpretable architectures such as Self-Explainable Neural Networks~\citep{alvarez2018towards} and Concept Whitening~\cite{chen2020concept}. However, our experiments show that all existing CBMs are limited to significant accuracy-vs-interpretability trade-offs. In this respect, our work reconciles theoretical results with empirical observations: while theoretical results suggest that explicit per-concept supervisions should improve generalisation bounds~\citep{li2018deep}, in contrast~\citet{koh2020concept}, \citet{chen2020concept}, and \citet{mahinpei2021promises} empirically show how learning with intermediate concepts may impair task performance in practice. The Information Plane method~\cite{shwartz2017opening} reveals that the higher generalisation error of existing concept bottleneck models might be explained as a compression in the input information flow caused by narrow architectures of Boolean and Fuzzy CBMs. In contrast, CEM represents the first concept-based model which does not need to compromise between task accuracy, concept interpretability or intervention power, thus filling this gap in the literature. Furthermore, through an ablation study on CEM's embedding size shown in Appendix~\ref{sec:appendix_emb_size_ablation},  we see that one does not need to increase the embedding size drastically to begin to see the benefits of using a CEM over a vanilla CBM or an end-to-end black box DNN. We note that as stronger interpretable models are deployed, there are risks of societal harm which we must be vigilant to avoid.


\paragraph{Conclusion}
Our experiments provide significant evidence in favour of CEM's accuracy/interpretability and, consequently, in favour of its real-world deployment. In particular, CEMs offer: (i) state-of-the-art task accuracy, (ii) interpretable concept representations aligned with human ground truths, (iii) effective interventions on learnt concepts, and (iv) robustness to incorrect concept interventions. While in practice CBMs require carefully selected concept annotations during training, which can be as expensive as task labels to obtain, our results suggest that CEM is more efficient in concept-incomplete settings, requiring less concept annotations and being more applicable to real-world tasks.
While there is room for improvement in both concept alignment and task accuracy in challenging benchmarks such as CUB or CelebA, as well as in resource utilisation during inference/training \reb{(see Appendix~\ref{sec:appendix_computational_time})}, our results indicate that CEM advances the state-of-the-art for the
accuracy-vs-interpretability trade-off, making progress on a crucial concern in explainable AI.

\begin{ack}
The authors would like to thank Carl Henrik Ek, Alberto Tonda, Andrei Margeloiu, Fabrizio Silvestri, and Maria Sofia Bucarelli for their insightful comments on earlier versions of this manuscript. MEZ acknowledges support from the Gates Cambridge Trust via a Gates Cambridge Scholarship. PB acknowledges support from the European Union's Horizon 2020 research and innovation programme under grant agreement No 848077. GC and FP acknowledges support from the EU Horizon 2020 project AI4Media, under contract no. 951911 and by the French government,
through Investments in the Future projects managed by the National Research Agency (ANR), 3IA Cote d’Azur with the reference number ANR-19-P3IA-0002. AW acknowledges support from a Turing AI Fellowship under grant EP/V025279/1, The Alan Turing Institute, and the Leverhulme Trust via CFI. GM is funded by the Research Foundation-Flanders (FWO-Vlaanderen, GA No 1239422N). FG is supported by TAILOR, a project funded by EU Horizon 2020 research and innovation programme under GA No 952215. This work was also partially supported by HumanE-AI-Net a project funded by EU Horizon 2020 research and innovation programme under GA 952026. MJ is supported by the EPSRC grant EP/T019603/1.
\end{ack}

\bibliographystyle{unsrtnat}
\bibliography{references}

\newpage

\section*{Checklist}


\begin{enumerate}

\item For all authors...
\begin{enumerate}
  \item Do the main claims made in the abstract and introduction accurately reflect the paper's contributions and scope?
    \answerYes{Section~\ref{sec:nesy_emb} and Section~\ref{sec:experiments} reflect our paper's contributions and scope as indicated in our introduction}
  \item Did you describe the limitations of your work?
    \answerYes{See our conclusion in Section~\ref{sec:discussion} for a discussion on some of the limitations of our approach.}
  \item Did you discuss any potential negative societal impacts of your work?
    \answerYes{} We discuss this in Section~\ref{sec:discussion}.
  \item Have you read the ethics review guidelines and ensured that your paper conforms to them?
    \answerYes{We have read the ethics review guidelines and ensured that your paper conforms to them.}
\end{enumerate}

\item If you are including theoretical results...
\begin{enumerate}
  \item Did you state the full set of assumptions of all theoretical results?
    \answerNA{}
        \item Did you include complete proofs of all theoretical results?
    \answerNA{}
\end{enumerate}

\item If you ran experiments...
\begin{enumerate}
  \item Did you include the code, data, and instructions needed to reproduce the main experimental results (either in the supplemental material or as a URL)?
    \answerYes{We uploaded a zip file with our code and documentation in the supplemental material and made our code available in a public repository\footnote{\href{https://github.com/mateoespinosa/cem/}{https://github.com/mateoespinosa/cem/}}.}
  \item Did you specify all the training details (e.g., data splits, hyperparameters, how they were chosen)?
    \answerYes{All training details are provided in Appendix~\ref{sec:appendix_architectures}
    }
        \item Did you report error bars (e.g., w.r.t. the random seed after running experiments multiple times)?
    \answerYes{Error bars are reported for all metrics and represent 95\% confidence intervals of the mean, unless specified otherwise.}
        \item Did you include the total amount of compute and the type of resources used (e.g., type of GPUs, internal cluster, or cloud provider)?
    \answerYes{Computational resources we used are specified in Appendix~\ref{sec:appendix_code}
    }
\end{enumerate}

\item If you are using existing assets (e.g., code, data, models) or curating/releasing new assets...
\begin{enumerate}
  \item If your work uses existing assets, did you cite the creators?
    \answerYes{References we used are listed in Appendix~\ref{sec:appendix_code} (code), Appendix~\ref{sec:appendix_architectures}
    (models), and Appendix~\ref{sec:appendix_data_details} (data).}
  \item Did you mention the license of the assets?
    \answerYes{We specified the licence of our new assets in Appendix~\ref{sec:appendix_code}.
    }
  \item Did you include any new assets either in the supplemental material or as a URL?
    \answerYes{We included our new assets (code, data, and models) in the supplemental material.}
  \item Did you discuss whether and how consent was obtained from people whose data you're using/curating?
    \answerNA{All new datasets are synthetic.}
  \item Did you discuss whether the data you are using/curating contains personally identifiable information or offensive content?
    \answerNA{All new datasets are synthetic.}
\end{enumerate}

\item If you used crowdsourcing or conducted research with human subjects...
\begin{enumerate}
  \item Did you include the full text of instructions given to participants and screenshots, if applicable?
    \answerNA{All new datasets are synthetic.}
  \item Did you describe any potential participant risks, with links to Institutional Review Board (IRB) approvals, if applicable?
    \answerNA{All new datasets are synthetic.}
  \item Did you include the estimated hourly wage paid to participants and the total amount spent on participant compensation?
    \answerNA{All new datasets are synthetic.}
\end{enumerate}

\end{enumerate}


\newpage
\appendix
\renewcommand\thefigure{\thesection.\arabic{figure}}

\section{Appendix}
\label{sec:appendix}

\subsection{Concept Alignment Score Implementation Details}
\label{sec:appendix_cas}
\reb{
As discussed in Section~\ref{sec:metrics}, there is lack of agreed-upon metrics to use for evaluating the interpretability of concept-based XAI models. For example, while concept predictive accuracy is well defined for scalar concept representations (e.g., vanilla CBMs), there seems to be no clear metric for evaluating the ``concept accuracy'' of an embedding representation. Therefore, in this work we build upon this gap and propose the CAS score as a generalization of the concept predictive accuracy. Intuitively, if a concept representation is able to capture a concept correctly, then we would expect that clustering samples based on that representation would result in coherent clusters where samples within the same cluster all have the concept active or inactive. The CAS attempts to capture this by looking at how coherent clusters are for each concept representation using the known concept labels for each sample as we change the size of each cluster. This is formally computed via Equation~\ref{eq:cas} throught a repeated evaluation of Rosenberg et al.'s homogeneity score~\cite{rosenberg2007v} for different clusterings.
}

Following~\citet{rosenberg2007v}, we compute the homogeneity score as described in Section~\ref{sec:metrics} by estimating the conditional entropy of ground truth concept labels $C_i$ w.r.t. cluster labels $\Pi_i$, i.e. $H(C_i,\Pi_i)$, using a contingency table. This table is produced by our selected clustering algorithm $\kappa$, i.e. $A=\{a_{u, v}\}$ where $a_{u, v}$ is the number of data points that are members of class $c_i = v \in \{0,1\}$ and elements of cluster $\pi_i = u \in \{1, \cdots, \rho\}$:
\begin{equation}
    H(C_i,\Pi_i) = - \frac{\rho}{N} \sum_{u=1}^\rho \Bigg( a_{u, 0} \log \frac{a_{u, 0}}{a_{u, 0} + a_{u, 1}} + a_{u, 1} \log \frac{a_{u, 1}}{a_{u, 0} + a_{u, 1}} \Bigg)
\end{equation}
Similarly, we compute the entropy of the ground truth concept labels $C_i$, i.e. $H(C)$, as:
\begin{equation}
    H(C_i) = - \Bigg( \frac{\sum_{u=1}^\rho a_{u, 0}}{2} \log \frac{\sum_{u=1}^\rho a_{u, 0}}{2} + \frac{\sum_{u=1}^\rho a_{u, 1}}{2} \log \frac{\sum_{u=1}^\rho a_{u, 1}}{2} \Bigg)
\end{equation}
When evaluating the CAS, we use $\delta = 50$ to speed up its computation across all datasets.

\subsection{Kernel Density Estimation of Mutual Information}
\label{sec:appendix_mi}
Following the approach of~\cite{kolchinsky2019nonlinear, michael2018on} we approximate the Mutual Information (MI) through the Kernel Density Estimation (KDE) method. ~\citet{kolchinsky2019nonlinear} show that this method accurately approximates the MI computed through the binning procedure proposed by~\citet{tishby2000information}. The KDE approach assumes that the activity of the analysed layer (in this case, the concept encoding layer $\hat{C}$) is distributed as a mixture of Gaussians.
This approximation holds true if the input samples used for evaluation are representative of the true input distribution.
Therefore, we can consider the input distribution as delta functions over each sample in the dataset. Moreover, Gaussian noise is added to the layer activity to bound the mutual information w.r.t. the input -- i.e., $\hat{C} = \mathbf{\hat{c}} + \epsilon$, where $\mathbf{\hat{c}}$ is the bottleneck activation vector and $\epsilon \sim N(0,\sigma^2I)$ is a noise matrix with noise variance $\sigma^2$. In this setting, the KDE estimation of the MI with the input is:

\begin{equation}
    I(\hat{C};X) = H(\hat{C}) - H(\hat{C}|X) = H(\hat{C}) \leq \frac{\zeta}{2} - \frac{1}{n} \sum_{i=1}^n\log\left(\frac{1}{n}\frac{1}{2\pi\sigma^2}\sum_{j=1}^n e^\frac{||\hat{c}^{(i)} - \hat{c}^{(j)}||_2^2}{2\sigma^2}\right),
\end{equation}
where $n$ is the number of input samples and $\zeta$ is the dimension of the concept encoding layer $\hat{C}$ (e.g., $\zeta =m \cdot k$ for CEM). Notice that ~\citet{shwartz2017opening} neglect the conditional entropy term arguing that the output of any neural network layer is a deterministic function of the input, which implies $H(\hat{C}|X) = 0$.

When considering instead the mutual information w.r.t. the downstream task label distribution $Y$, the conditional entropy is  $H(\hat{C}|Y)\neq0$ and the mutual information $I(\hat{C};Y)$ can be estimated as:

\begin{equation*}
\begin{split}
    I(\hat{C};Y) = H(\hat{C}) - H(\hat{C}|Y) 
    &\leq \frac{\zeta}{2} - \frac{1}{n} \sum_{i=1}^n\log\left(\frac{1}{n}\frac{1}{2\pi\sigma^2}\sum_{j=1}^n e^\frac{||\hat{c}^{(i)} - \hat{c}^{(j)}||_2^2}{2\sigma^2}\right)\\ 
    &- \sum^L_{l=1} p_l\left[\frac{\zeta}{2}-\frac{1}{P_l}\sum_{\substack{i \\ \text{s.t. } y^{(i)} = l}}\log \left(\frac{1}{P_l}\frac{1}{2\pi\sigma^2}\sum_{\substack{j \\ \text{s.t. } y^{(j)} = l}}e^\frac{||\hat{c}^{(i)} - \hat{c}^{(j)}||_2^2}{2\sigma^2}\right)\right],
\end{split}
\end{equation*}

where $L$ is the number of downstream task labels, $P_l$ the number of data with output label $l$, and $p_l = P_l/n$ is the probability of task label $l$. 

When considering the concept labels $C$, however, the same estimation cannot be employed since it requires the labels to be mutually exclusive. While this holds true for the task labels $Y$ in the considered settings, the concepts in $C$ are generally multi-labeled — i.e., more than one concept can be true when considering a single sample $\mathbf{x}^{(i)}$. Therefore, in this case we compute the average of the conditional entropies $H(\hat{C}|C) = 1/k\sum_a H(\hat{C}|C_a)$ across all $k$ concepts. More precisely,

\begin{equation*}
\begin{split}
    I(\hat{C};C) &= H(\hat{C}) - H(\hat{C}|C) \\
    & = H(\hat{C}) - \frac{1}{k}\sum_{a=1}^k H(\hat{C}|C_a)\\
    &\leq \frac{\zeta}{2} - \frac{1}{n} \sum_{i=1}^n\log\left(\frac{1}{n}\frac{1}{2\pi\sigma^2}\sum_{j=1}^n e^\frac{||\mathbf{\hat{c}}^{(i)} - \mathbf{\hat{c}}^{(j)}||_2^2}{2\sigma^2}\right)\\ 
    &- \frac{1}{k}\sum^k_{a=1} \sum_{m\in M_a} p_{a,m} 
    \left[\frac{\zeta}{2}-\frac{1}{P_{a,m}}\sum_{\substack{i \\ \text{s.t. } c_{a}^{(i)} = m}} \log \left(\frac{1}{P_{a,m}}\frac{1}{2\pi\sigma^2}\sum_{\substack{j \\ \text{s.t. } c_{a}^{(j)} = m}} e^\frac{||\mathbf{\hat{c}}^{(i)} - \mathbf{\hat{c}}^{(j)} ||_2^2}{2\sigma^2}\right)\right],
\end{split}
\end{equation*}

where $P_{a,m}$ is the number of samples having the concept $c_a = m$, $M_k$ is the set of possible values that the $c_a$ concept can assume (generally $M_a = \{0,1\}$), and $p_{a,m} = P_{a,m}/n$ is the probability of concept label $c_a = m$. 

In all the previous cases, since we use the natural logarithm, the MI is computed in NATS. To convert it into bits, we scale the obtained values by $\frac{1}{\log(2)}$.

\paragraph{The role of noise}
The variance $\sigma^2$ of the noise matrix $\epsilon$, plays an important role in the computation of the MI. More precisely, low values of $\sigma$ entail high negative values for $H(\hat{C}|X)$, and, consequently, high positive values for $I(\hat{C};X)$. In the extreme case where we do not add any noise, we have $H(\hat{C}|X) = -\inf$ and $I(\hat{C};X) \sim \inf$, as long as the entropy $H(\hat{C})$ is finite. 
Furthermore, as we can observe in the equations above, the dimensionality $\zeta$ of the concept representation also plays an important role in the computation of the MI, the latter being directly proportional to the dimensionality of concept representation layer $\hat{C}$. To mitigate this issue, we also consider the noise to be directly proportional to the dimension of $\hat{C}$, by setting $\sigma^2 = \zeta/100$.

\subsection{Datasets}
\label{sec:appendix_data_details}

\subsubsection{XOR problem}
The first dataset used in our experiments is inspired by the exclusive-OR (XOR) problem proposed by~\cite{minsky2017perceptrons} to show the limitations of Perceptrons. We draw input samples from a uniform distribution in the unit square $\mathbf{x} \in [0,1]^2$ and define two binary concepts \{$c_1, c_2\}$ by using the Boolean (discrete) version of the input features $c_i = \mathbb{1}_{x_i > 0.5}$. Finally, we construct a downstream task label using the XOR of the two concepts $y = c_1 \oplus c_2$.


\subsubsection{Trigonometric dataset}
The second dataset we use in our experiments is inspired by that proposed by ~\citet{mahinpei2021promises} (see Appendix D of their paper). Specifically, we construct synthetic concept-annotated samples from three independent latent normal random variables $h_i \sim \mathcal{N}(0, 2)$. Each of the 7 features in each sample is constructed via a non-invertible function transformation of the latent factors, where 3 features are of the form $(\sin(h_i) + h_i)$, 3 features of the form $(\cos(h_i) + h_i)$, and 1 is the nonlinear combination $(h_1^2 + h_2^2 + h_3^2)$. Each sample is then associated with 3 binary concepts representing the sign of their corresponding latent variables, i.e. $c_i = (h_i>0)$. In order to make this task Boolean-undecidable from its binary concepts, we modify the downstream task proposed by~\citet{mahinpei2021promises} by assigning each sample a label $y = \mathbb{1}_{(h_1 + h_2) > 0}$ indicating whether $h_1 + h_2$ is positive or not.


\subsubsection{Dot dataset} \label{sec:appendix_dot_dataset}
As much as the Trigonometric dataset is designed to highlight that fuzzy concept representations generalize better than Boolean concept representations, we designed the Dot dataset to show the advantage of embedding concept representations over fuzzy concept representations. The Dot dataset is based on four 2-dimensional latent factors from which concepts and task labels are constructed. Two of these four vectors correspond to fixed reference vectors $\mathbf{w}_+$ and $\mathbf{w}_-$ while the remaining two vectors $\{\textbf{v}_i\}_{i=1}^2$ are sampled from a 2-dimensional normal distribution:
\begin{align}
    \mathbf{v}_{1,2} \sim \mathcal{N}(\mathbf{0}, 2\ \mathbf{I}) \qquad
    \mathbf{w}_+ = 
    \begin{bmatrix}
        1 & 1
    \end{bmatrix}^T \qquad
    \mathbf{w}_- = -\mathbf{w}_+
\end{align}
We then create four input features as the sum and difference of the two factors $\mathbf{v}_i$:
\begin{align}
    \mathbf{x} = 
    \begin{bmatrix}
        (\mathbf{v}_1 + \mathbf{v}_2) & (\mathbf{v}_1 - \mathbf{v}_2)
    \end{bmatrix}^T
\end{align} 
From this, we create two binary concepts representing whether or not the latent factors $\mathbf{v}_i$ point in the same direction as the reference vectors $\mathbf{w}_j$ (as determined by their dot products):
\begin{equation}
    \mathbf{c} = 
    \begin{bmatrix}
        \mathbb{1}_{(\mathbf{v}_1 \cdot \mathbf{w}_1) > 0} & \mathbb{1}_{(\mathbf{v}_2 \cdot \mathbf{w}_2) > 0}
    \end{bmatrix}^T
\end{equation}

Finally, we construct the downstream task as determining whether or not vectors $\mathbf{v}_1$ and $\mathbf{v}_2$ point in the same direction (as determined by their dot product):

\begin{equation}
    y = \mathbb{1}_{(\mathbf{v}_1 \cdot \mathbf{v}_2) > 0}
\end{equation}

\subsubsection{Real-world datasets} \label{sec:appendix_vision_datasets}
Furthermore, we evaluate our methods on two real-world vision tasks: (1) the Caltech-UCSD Birds-200-2011 dataset (CUB,~\citep{wah2011caltech}), as prepared by~\citep{koh2020concept}, and the Large-scale CelebFaces Attributes dataset (CelebA,~\citep{liu2015deep}).

\paragraph{CUB~\cite{wah2011caltech}} In CUB we construct a dataset with complete concept annotations by using the same $k = 112$ bird attributes selected by~\citet{koh2020concept} as binary concept annotations (e.g., \textit{beak\_type}, \textit{wing\_color}, etc ...) and using the bird identity ($l = 200$) as the downstream task. All images are preprocessed in the same fashion as in~\cite{koh2020concept} by normalizing and randomly flipping and cropping each image during training. This results in a dataset of around 6,000 RGB images with sizes $(3, 299, 299)$ which are split into test, validation, and training sets using the same splits by~\citet{koh2020concept}. In our evaluation, we use CUB to test CBMs in real-world tasks where we have a complete set of concept annotations w.r.t. the downstream task.

\paragraph{CelebA~\cite{liu2015deep}} In CelebA, we select the $8$ most balanced attributes $[a_1, \cdots a_8]$ out of each image's $40$ binary attributes, as defined by how close their distributions are to a random uniform binary distribution, and use attributes $[a_1, \cdots, a_6]$ as concepts annotations for each sample. To simulate a task in which complete concept annotations are lacking, each image in CelebA is assigned a label corresponding to the base-10 representation of the number formed by the binary vector $[a_1, \cdots, a_8]$, resulting in a total of $l = 2^8 = 256$ classes. Note that concept annotations in this task are incomplete as attributes $a_7$ and $a_8$ are needed for predicting the downstream task but they are not provided during training. To improve resource utilization and training times, we further reduce the size of the CelebA dataset by randomly subsampling the dataset and selecting every $12^\text{th}$ sample during training and we downsample every image to have shape $(3, 64, 64)$. This results in a dataset with around 16,900 RGB images from which a train, validation, and test datasets are generated using a traditional $70\%$-$10\%$-$20\%$ split.  In our experiments, we use CelebA to evaluate CBMs in scenarios where the bottleneck is extremely narrow and incomplete w.r.t. the downstream task.

\subsection{Effect of Concept Encoder Capacity}
\label{sec:appendix_resnet18}
\begin{figure}[h!]
    \centering
    \includegraphics[width=0.8\textwidth]{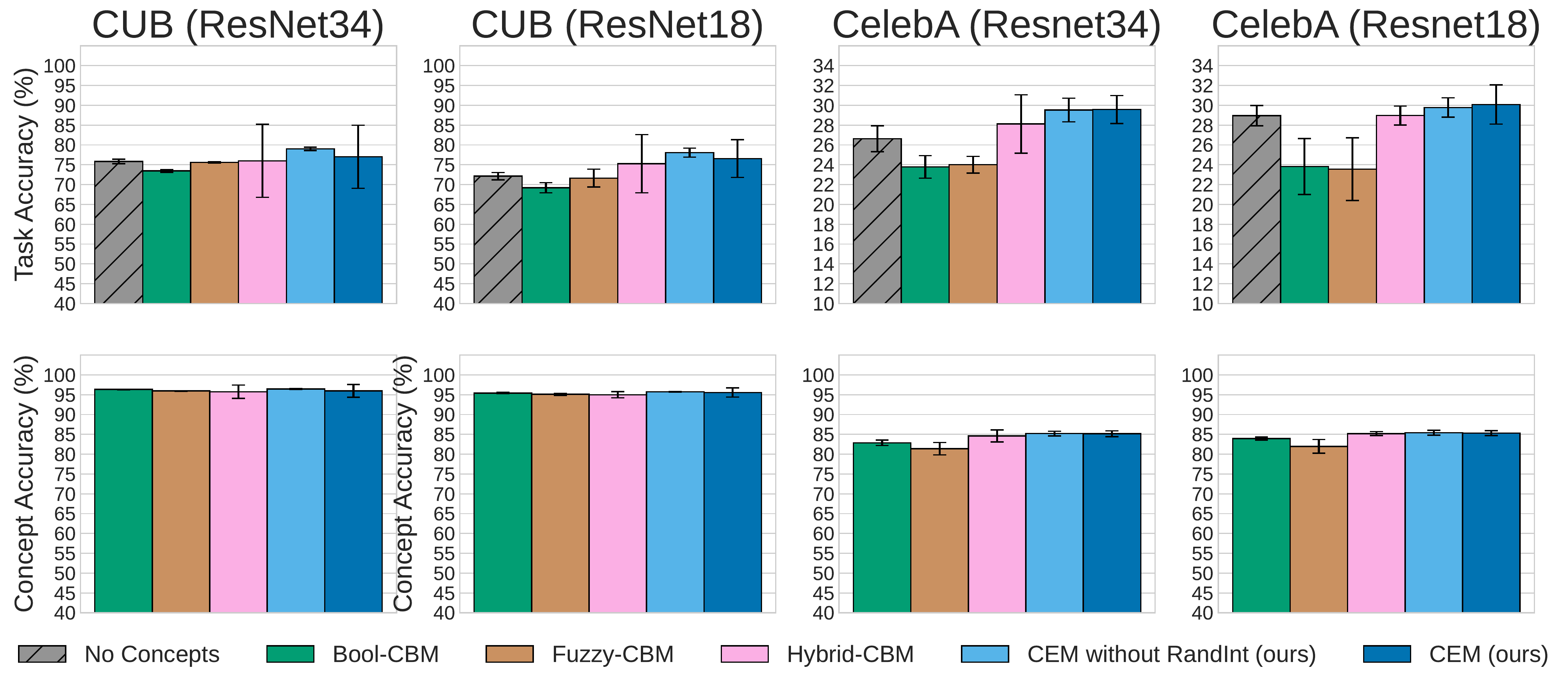}
    \caption{Task and average concept accuracy when using a ResNet18 backbone vs a ResNet34 backbone in CUB and CelebA.}
    \label{fig:backbone_perfs}
\end{figure}

\reb{
Different concept encoders will have different approximation capabilities, and the resulting concept representations will be affected by the architectural choices. To test whether the choice of a specific model might bias our results, here we show that the relative rankings across methods in our real-world tasks (CUB and CelebA) are preserved when using backbones with significantly different capacities i.e., a ResNet18 vs a ResNet34. Specifically, Figure~\ref{fig:backbone_perfs} compares the concept and task predictive accuracies of our baselines in CUB and CelebA when using different backbone capacities (trained while fixing all other hyperparamters are described in Appendix~\ref{sec:appendix_architectures}). Notice that although we observe a drop in performance when using a ResNet18 backbone, this drop is similar across all baselines and therefore leads to our results having the same ranking as observed when using a ResNet34 backbone.}

\begin{figure}[h!]
    \centering
    \includegraphics[width=0.9\textwidth]{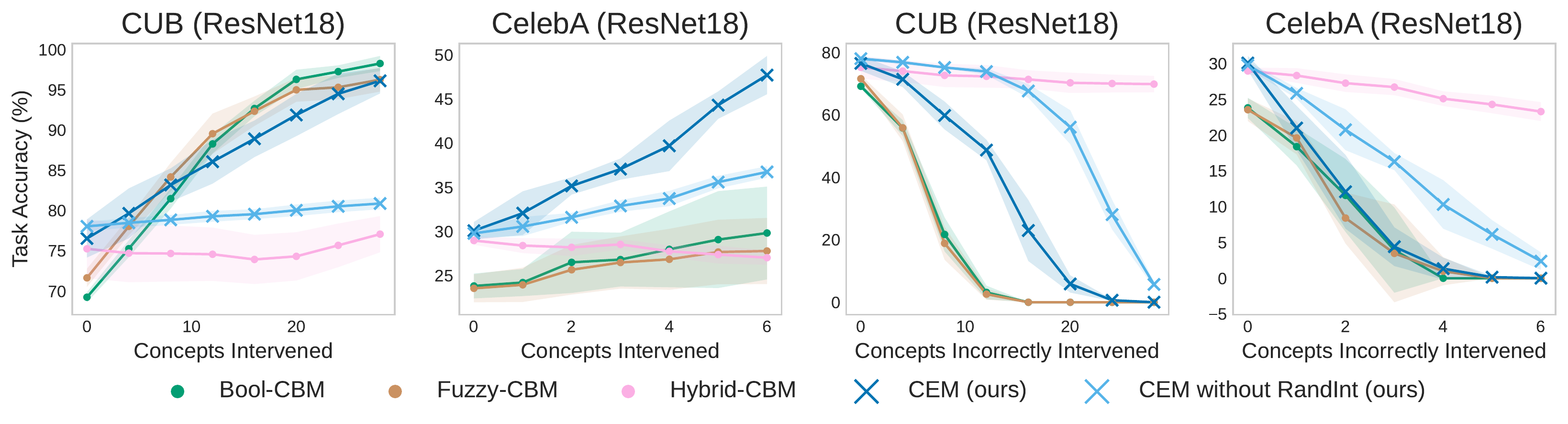}
    \caption{Effects of performing positive random concept interventions (left and center left) and incorrect random interventions (center right and right) for different models with a ResNet18 backbone in CUB and CelebA. As in~\citep{koh2020concept}, when intervening in CUB we jointly set groups of mutually exclusive concepts.}
    \label{fig:backbone_int}
\end{figure}

\reb{Similarly, Figure~\ref{fig:backbone_int} shows that the same rankings and results observed in Figure~\ref{fig:interventions}, where a ResNet34 backbone was used, can be seen when performing interventions in the baselines which use a ResNet18 backbone.}

\subsection{RandInt Probability Ablation Study}
\label{sec:appendix_prob_ablation}
\begin{figure}[h!]
    \centering
    \begin{subfigure}[b]{0.5\textwidth}
        \includegraphics[width=\textwidth]{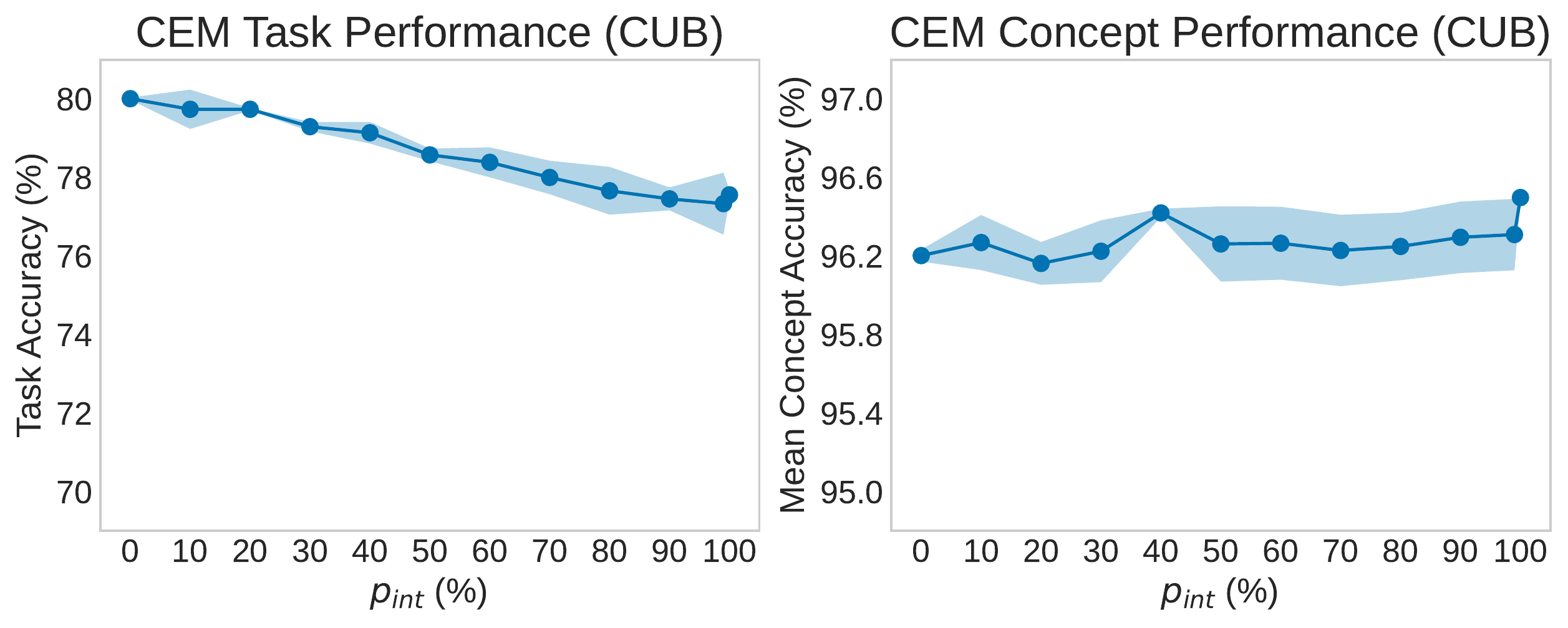}
        \subcaption{}
    \end{subfigure}
    \begin{subfigure}[b]{0.35\textwidth}
        \includegraphics[width=\textwidth]{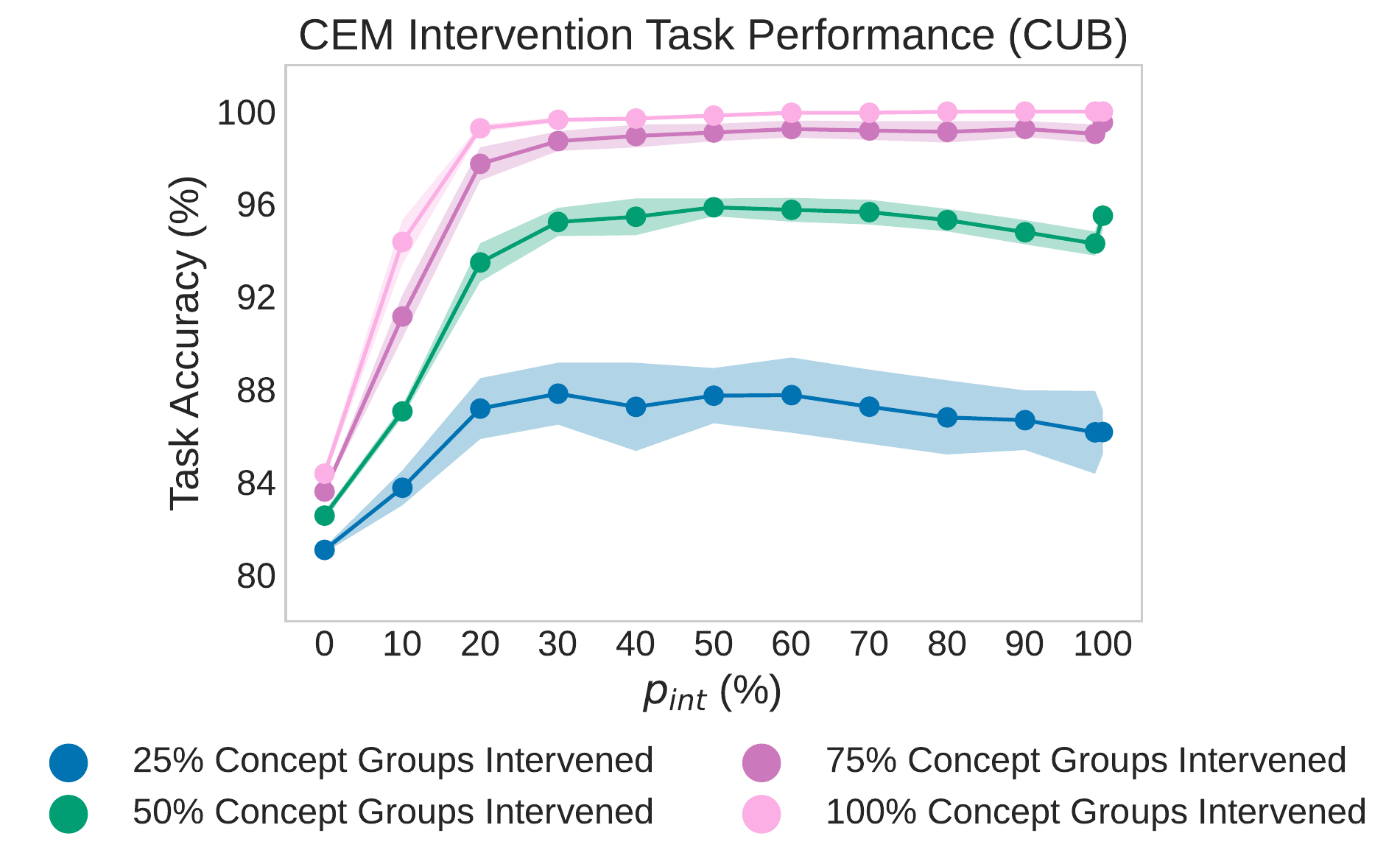}
        \subcaption{}
    \end{subfigure}
    \caption{Ablation study for $p_\text{int}$ in CUB. (a) Task and concept validation accuracy of CEMs trained with different values of $p_\text{int}$. (b) Task validation accuracy when intervening on an increasing number of concept groups for CEMs trained with different values of $p_\text{int}$.}
    \label{fig:prob_ablation}
\end{figure}

Figure~\ref{fig:prob_ablation} shows the results of varying $p_\text{int}$ for CEMs trained on CUB (using the same training setup as defined in Appendix~\ref{sec:appendix_architectures}). We observe that although there is a slight trade-off in validation task accuracy as we increase $p_\text{int}$, this trade-off is eclipsed compared to the concept intervention capabilities which come by increasing $p_\text{int}$. Because of this, in our work we settle with $p_\text{int} = 0.25$ as this study shows that this value leverages good performance without interventions while enabling effective interventions.

\subsection{Training Details} \label{sec:appendix_architectures}
\paragraph{Model Architectures} For simplicity, we use the same DNN architectures across all synthetic tasks (i.e., \textit{XOR}, \textit{Trig}, \textit{Dot}) unless specified otherwise. Specifically, we use an MLP with hidden layer sizes $\{128, 128\}$ and LeakyReLU activations for latent code generator $\psi$ in CEM and concept encoder $g$ in all CBM variants. When learning concept embedding representations in synthetic datasets, we learn embeddings with $m=128$ activations.

In both CUB and CelebA, for latent code generator $\psi$ in CEM and concept encoder $g$ in all CBM variants we use a pretrained ResNet-34 model~\cite{he2016deep} with its last layer modified to output $n_\text{hidden} = m$ activations. When using CEM, we learn embeddings with $m=16$ activations, smaller than in the synthetic datasets given the larger number of concepts in these tasks (see Appendix~\ref{sec:appendix_emb_size_ablation} for an ablation study showing how the embedding size affects performance in CEM).

Across all datasets we always use a single fully connected layer for label predictor $f$ and, for the sake of fairness, set $\gamma = k\cdot (m - 1)$ when evaluating Hybrid CBMs.
\reb{This is done so that the overall bottleneck of Hybrid-CBM has size $k + \gamma = k + k (m-1) = k m$, just as in an equivalent CEM model. Notice therefore that the dimensionality of $\mathbf{\hat{c}}$ is $k$ for Bool and Fuzzy CBMs while it is $k \cdot m$ for our Hybrid-CBM and CEM baselines.}
When training end-to-end models without concept supervision (i.e., our ``No Concepts'' baseline), we use the exact same architecture as in the Hybrid-CBM but provide no concept supervision in its bottleneck (equivalent to setting the weight for the concept loss to $0$ during training). Finally, when using RandInt, we set  $p_\text{int} = 0.25$, as empirically we observe that this yields good results across all datasets (see Appendix~\ref{sec:appendix_prob_ablation} above).

\paragraph{Training Hyperparameters} In all synthetic tasks, we generate datasets with 3,000 samples and use a traditional 70\%-10\%-20\% random split for training, validation, and testing datasets, respectively. During training, we then set the weight of the concept loss to $\alpha = 1$ across all models. We then train all models for 500 epochs using a batch size of $256$ and a default Adam~\cite{kingma2014adam} optimizer with learning rate $10^{-2}$.

In CUB, we set the concept loss weight to $\alpha = 5$ in all models and, as in~\cite{koh2020concept}, we use a weighted cross entropy loss for concept prediction to mitigate imbalances in concept labels. All models in this task are trained for 300 epochs using a batch size of 128 and an SGD optimizer with $0.9$ momentum and learning rate of $10^{-2}$.

In our CelebA task, we fix the concept loss weight to $\alpha = 1$ in all models and also use a weighted cross entropy loss for concept prediction to mitigate imbalances in concept labels. All models in this task are trained for 200 epochs using a batch size of 512 and an SGD optimizer with $0.9$ momentum and learning rate of $5 \times 10^{-3}$ (different from CUB to avoid instabilities observed if the initial learning rate was too high).

In all models and tasks, we use a weight decay factor of $4e-05$ and scale the learning rate during training by a factor of $0.1$ if no improvement has been seen in validation loss for the last $10$ epochs. Furthermore, all models are trained using an early stopping mechanism monitoring validation loss and stopping training if no improvement has been seen for 15 epochs.

\subsection{Task and Mean Concept Performance} \label{sec:appendix_taks_and_concept_perf}
\begin{figure}[!h]
    \centering
    \begin{subfigure}[b]{0.8\textwidth}
        \includegraphics[width=\textwidth]{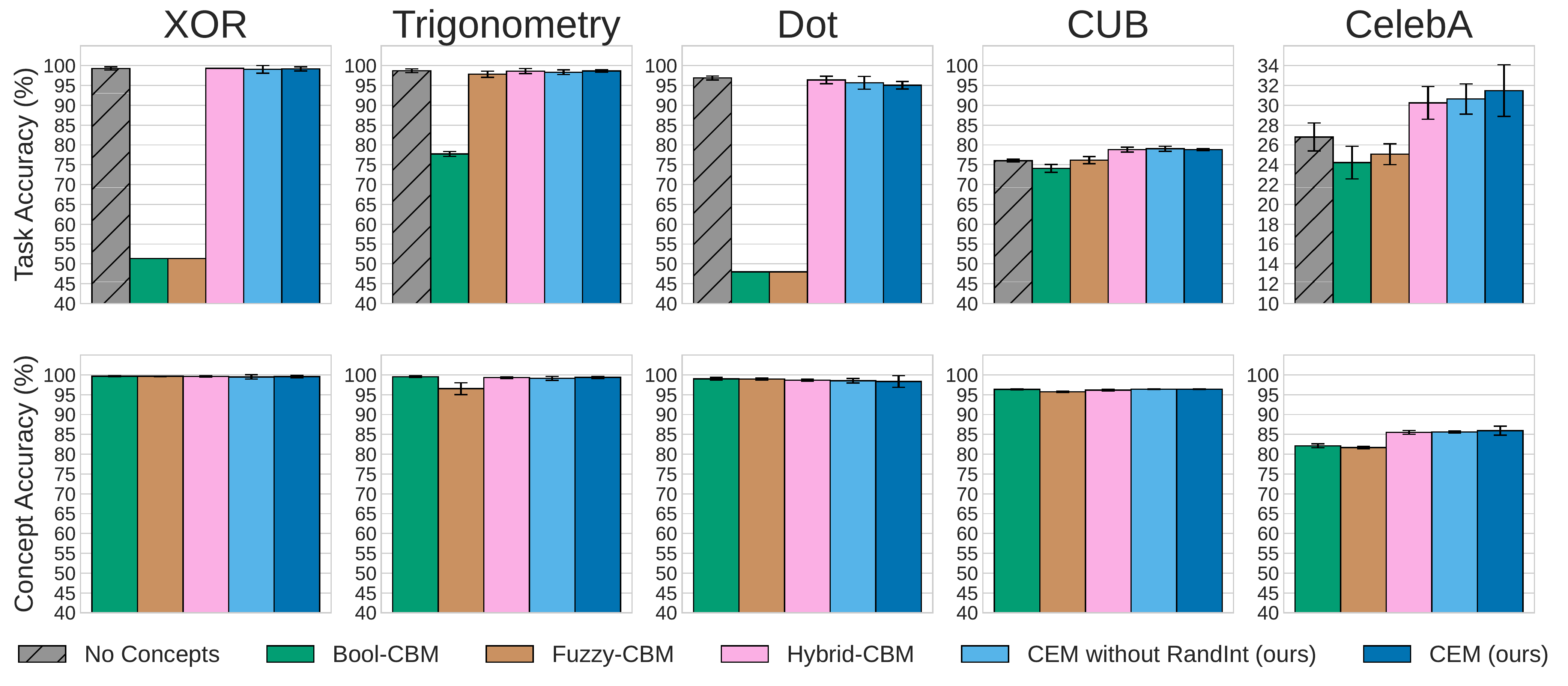}
        \subcaption{}
        \label{fig:task_and_perf}
    \end{subfigure}
    \begin{subfigure}[b]{0.8\textwidth}
        \centering
        \includegraphics[width=\textwidth]{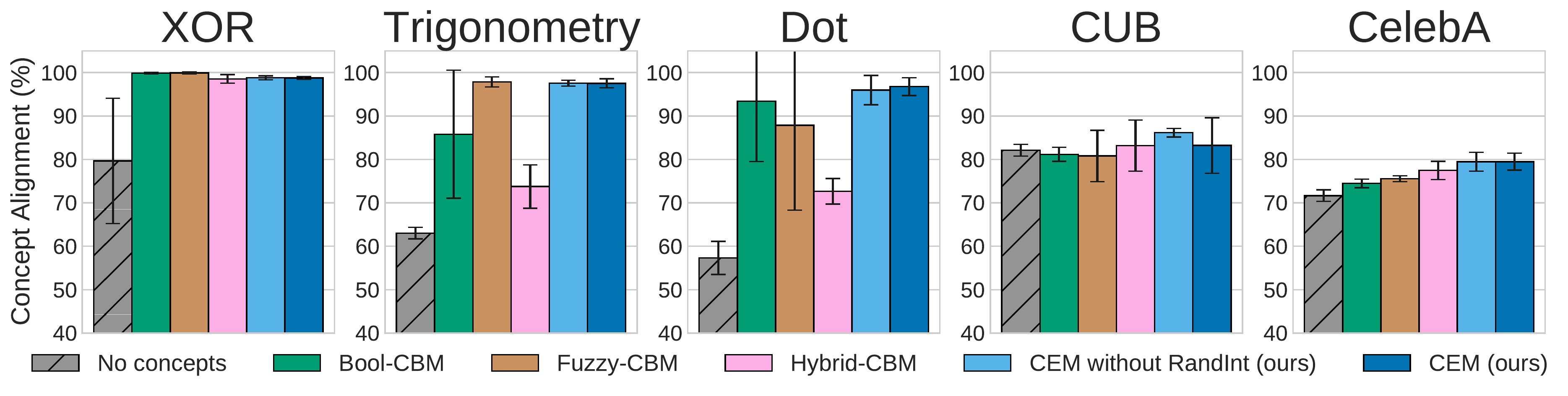}
        \subcaption{}
        \label{fig:alignment_scores}
    \end{subfigure}
    
    \caption{(a) Task and mean concept accuracy for all methods across all tasks. (b) Concept alignment scores for all methods across all tasks.}
\end{figure}

In Figure~\ref{fig:task_and_perf} we show the task and mean concept predictive performance of all of our baselines. Notice that as claimed in Section~\ref{sec:experiments}, all baselines are able to achieve a very similar mean concept accuracy but they have very distinct task accuracies, suggesting a that the interpretability-vs-accuracy trade-off is different across different models. For further clarity and to facilitate cross-comparison across methods and datasets, we also show our concept alignment scores in a bar-plot format in Figure~\ref{fig:alignment_scores}.

\reb{In Table~\ref{tab:cas} and Table~\ref{tab:accuracy} we report the same results in tabular form for clarity's sake. Notice how in CUB the baseline model without concept suppervision (i.e., ``No Concepts'') has a better CAS mean compared to Bool and Fuzzy CBMs. We hypothesize that because certain concepts in CUB tend to be activated only for specific classes (e..g, there is a very high imbalance in how concepts are activated across classes), clusters produced from the intermediate representations of a DNN trained to predict said classes will be highly coherent with respect to concepts that are class-specific, leading to high CAS scores. The same cannot be said of e.g., CelebA (where concept activations are highly balanced across different classes by design), which is why we observe the CAS in black-box DNNs being lower than that in CBM models.}

\begin{table}[h!]
\centering
\caption{\reb{Task accuracy for all methods across all tasks reported with the mean and 95\% confidence interval.}}
\resizebox{\textwidth}{!}{%
\begin{tabular}{llllll}
\toprule
 & No concepts & Boolean-CBM & Fuzzy-CBM & Hybrid-CBM & CEM (ours) \\ 
\midrule
\textbf{XOR} & \textbf{$99.33$, $(99.01, 99.66)$} & $51.33$, $(51.33, 51.33)$ & $51.42$, $(51.42, 51.42)$ & \textbf{$99.23$, $(99.23, 99.23)$} & \textbf{$99.17$, $(98.71, 99.57)$} \\
\textbf{Trigonometry} & \textbf{$98.47$, $(98.47, 98.47)$} & $77.77$, $(77.52, 77.99)$ & $98.37$, $(98.37, 98.37)$ & \textbf{$98.67$, $(98.42, 98.90)$} & \textbf{$98.43$, $(97.79, 99.01)$} \\
\textbf{Dot} & \textbf{$97.57$, $(97.01, 98.09)$} & $48.00$, $(48.00, 48.00)$ & $48.17$, $(48.02, 48.31)$ & $96.67$, $(96.67, 96.67)$ & \textbf{$97.13$, $(97.13, 97.13)$} \\
\textbf{CUB} & $73.41$, $(71.83, 74.70)$ & $67.11$, $(65.29, 68.56)$ & \textbf{$72.98$, $(70.39, 76.30)$} & \textbf{$70.70$, $(64.28, 77.68)$} & \textbf{$77.11$, $(75.89, 78.10)$} \\
\textbf{CelebA} & $26.80$, $(25.90, 27.84)$ & $24.23$, $(24.23, 24.23)$ & $25.07$, $(24.36, 25.81)$ & \textbf{$30.24$, $(29.13, 31.41)$} & \textbf{$30.63$, $(29.62, 31.74)$}\\
\bottomrule
\end{tabular}%
}
\label{tab:accuracy}
\end{table}
\begin{table}[h!]
\centering
\caption{\reb{Concept alignment scores for all methods across all tasks reported with the mean and 95\% confidence interval.}}
\resizebox{\textwidth}{!}{%
\begin{tabular}{llllll}
\toprule
 & No concepts & Boolean-CBM & Fuzzy-CBM & Hybrid-CBM & CEM (ours) \\ 
\midrule
\textbf{XOR} & $79.65$, $(71.32, 89.12)$ & $99.86$, $(99.86, 99.86)$ & \textbf{$99.92$, $(99.92, 99.92)$} & $98.53$, $(97.88, 99.10)$ & $98.79$, $(98.50, 99.06)$ \\
\textbf{Trigonometry} & $63.02$, $(62.18, 63.66)$ & $85.80$, $(85.80, 85.80)$ & \textbf{$97.84$, $(97.84, 97.84)$} & $73.75$, $(73.75, 73.75)$ & \textbf{$97.55$, $(97.11, 97.93)$} \\
\textbf{Dot} & $57.31$, $(53.80, 57.31)$ & \textbf{$93.40$, $(85.22, 99.57)$} & \textbf{$87.86$, $(75.03, 98.24)$} & $72.66$, $(70.68, 74.26)$ & \textbf{$95.98$, $(94.90, 97.16)$} \\
\textbf{CUB} & $82.12$, $(81.49, 82.69)$ & $81.18$, $(80.12, 82.09)$ & $80.79$, $(79.36, 82.75)$ & \textbf{$83.19$, $(79.81, 85.78)$} & \textbf{$86.14$, $(85.50, 86.68)$} \\
\textbf{CelebA} & $71.66$, $(71.66, 71.66)$ & $74.48$, $(73.87, 75.08)$ & $75.56$, $(75.16, 75.91)$ & $77.48$, $(77.48, 77.48)$ & \textbf{$79.47$, $(78.43, 80.33)$}\\
\bottomrule
\end{tabular}%
}
\label{tab:cas}
\end{table}

\subsection{Concept Subsampling in CUB}
\label{sec:appendix_concept_subsampling}
\begin{figure}[!h]
    \centering
    \includegraphics[width=0.5\textwidth]{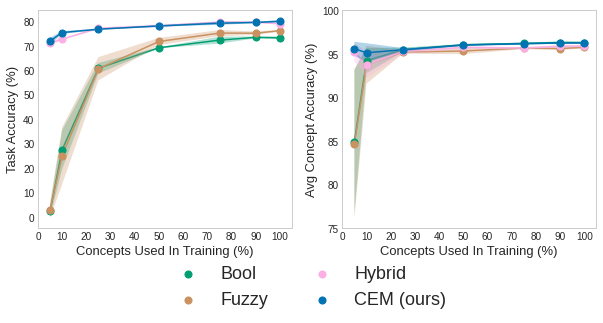}
    \caption{Task and average concept accuracies when using a percentage of the available concepts in our CUB task during training. All points are generated by sampling, uniformly at random, 5 different concept subsets at training time and averaging all metrics.}
    \label{fig:scarce-annotations}
\end{figure}

\reb{
All concept bottleneck models require datasets containing concept annotations, which may be costly to acquire. Here we compare a CBM's robustness when concept annotations are scarce. We simulate this scenario by randomly selecting a random subsample of the 112 concepts in our CUB task which we then use as annotations for all models during training (all models are trained using the same architecture and training hyperparameters as our CUB model in Section~\ref{sec:experiments}). As we observe in Figure~\ref{fig:scarce-annotations}, the task and concept accuracy of both CEMs and Hybrid-CBMs are only mildly affected by the reduction in concept supervisions, as opposed to Bool and Fuzzy CBMs. In both CEMs and Hybrid-CBMs this robustness allows a dramatic reduction of required concept annotations and the costs related to acquiring such annotations.  Nevertheless, as seen in Section~\ref{sec:int_results}, notice that although Hybrid-CBM performs well in concept scarsity, it is unable to effectively react to human concept interventions (a crucial limitation that CEM is able to overcome).
}

\subsection{Bottleneck Representation Experiment Details}
\label{sec:appendix_repr_power_experiment}
To explore our hypothesis that the high alignment observed in CEM's representations may lead to its embeddings forming more interpretable representations than Hybrid's embeddings, we evaluate the power of their learnt bottlenecks as representations for different tasks. With this aim, we train a Hybrid-CBM and a CEM, both with the same architecture as described for models trained on CUB in Appendix~\ref{sec:appendix_architectures}, on a variation of CUB with only 25\% of its concept annotations randomly selected before training. This results on a total of $k = 28$ concepts being randomly selected to be provided as supervision for both models. We then train these models to convergence using the same training setup as in CUB models described in Appendix~\ref{sec:appendix_architectures} resulting in a Hybrid-CBM with $77.15\% \pm 0.33\%$ test task accuracy and  $95.3\% \pm 0.31\%$ test mean concept accuracy. In contrast, its CEM counterpart achieved $76.76\% \pm 0.27\%$ test task accuracy and $95.47\% \pm 0.19\%$ test mean concept accuracy.

Once trained, we use the bottleneck representations learn by both the Hybrid-CBM and the CEM to predict the remaining 75\% of the concept annotations in CUB using a simple logistic linear model. In other words, for each concept not used to train each of these models (of which there are $112 - 28 = 84$ of them) we train a linear probe to predict the concept's true value from the entire bottleneck representations learnt by both our Hybrid-CBM and CEM models. We do this for a total of 5 randomly initialized Hybrid-CBMs and CEMs and observe that the probes trained using the Hybrid-CBM's bottleneck have a mean concept accuracy of 91.83\% $\pm$ 0.51\% while the probes trained using CEM's bottleneck have a mean concept accuracy of 94.33\% $\pm$ 0.88\%.

\subsection{More Qualitative Results}
\label{sec:appendix_qualitative}
\begin{figure}[h!]
    \centering
    \begin{subfigure}[b]{0.3\textwidth}
        \centering
        \includegraphics[width=\textwidth]{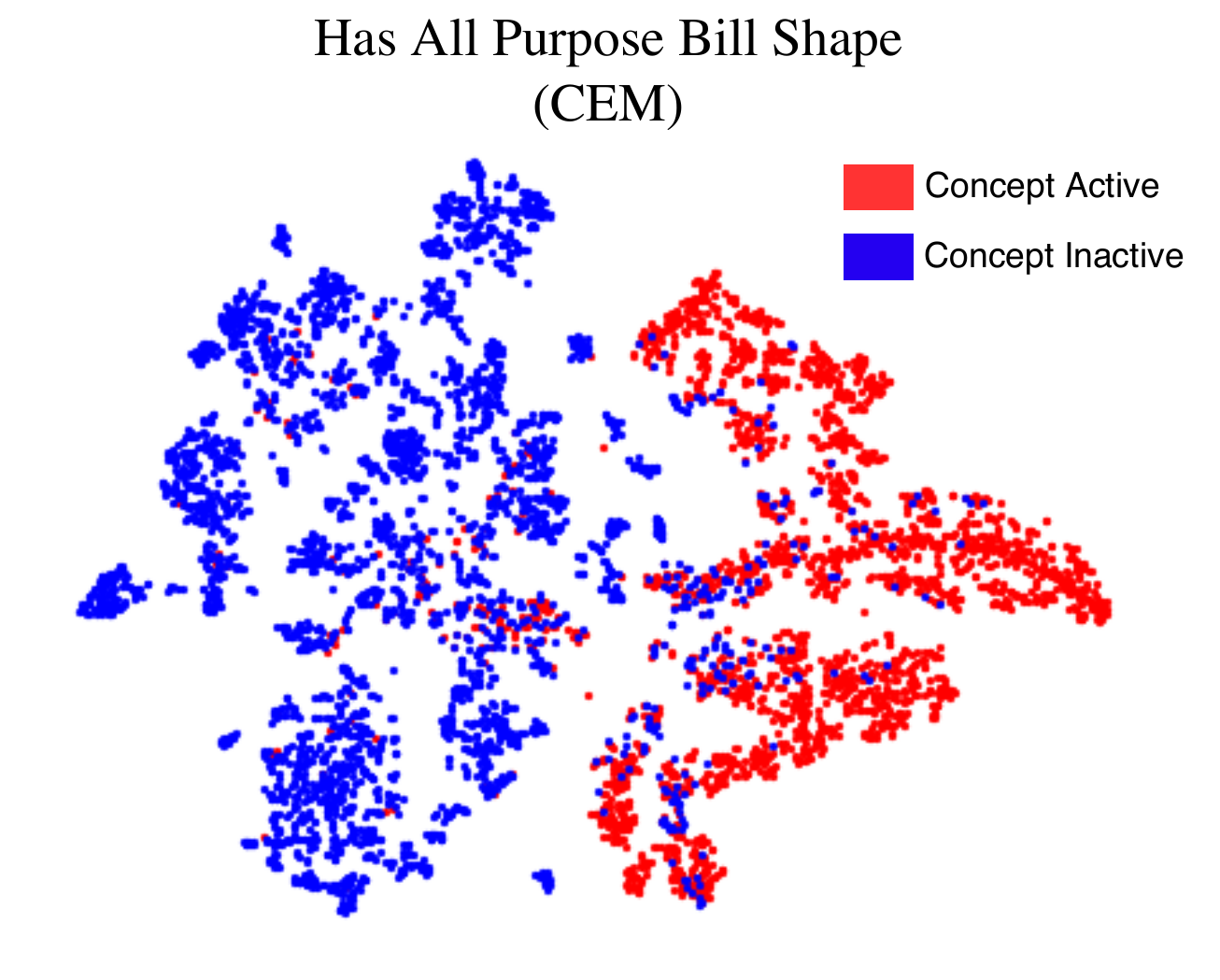}
        \includegraphics[width=\textwidth]{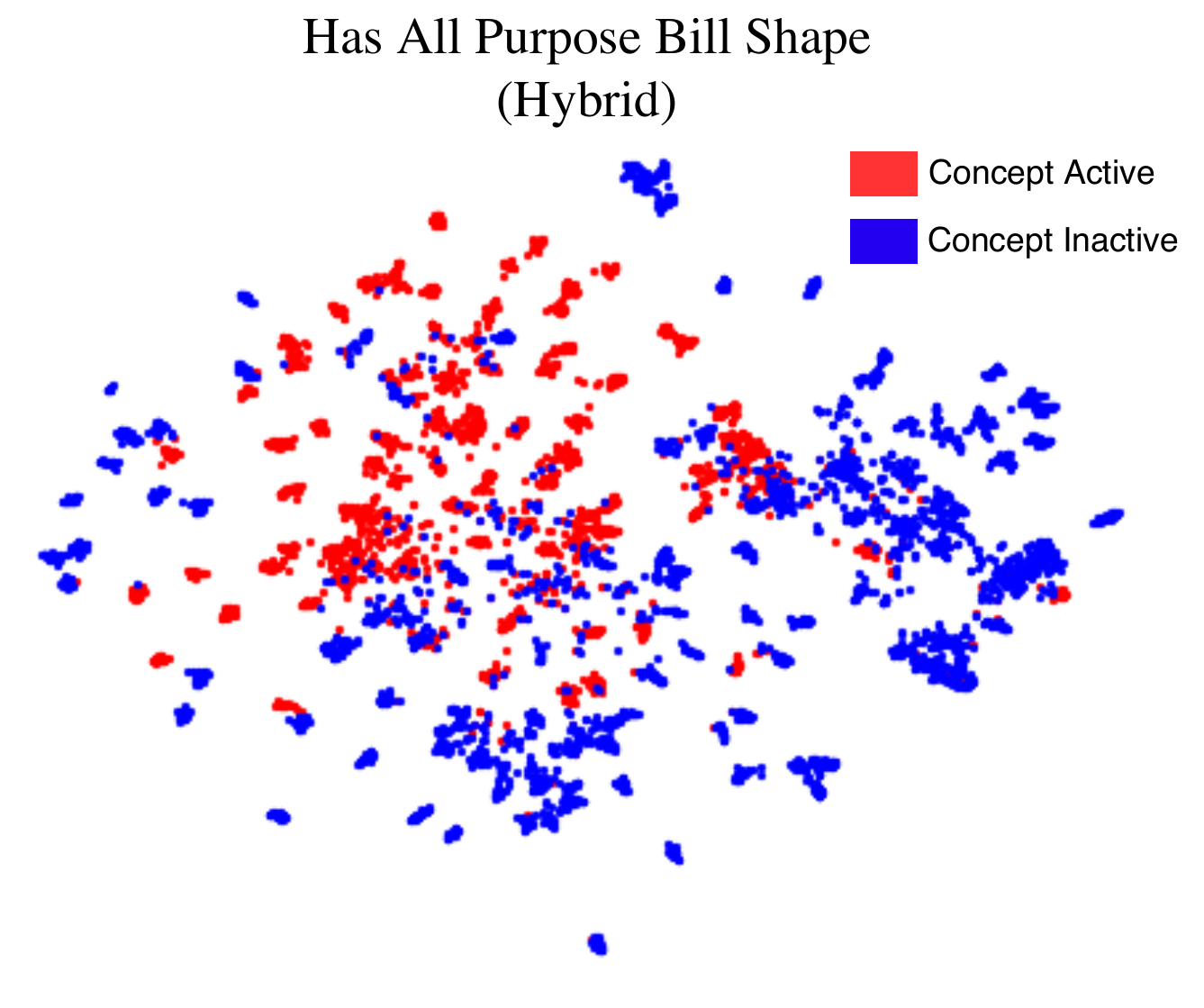}
        \subcaption{
        }
    \end{subfigure}
    \begin{subfigure}[b]{0.3\textwidth}
        \centering
        \includegraphics[width=\textwidth]{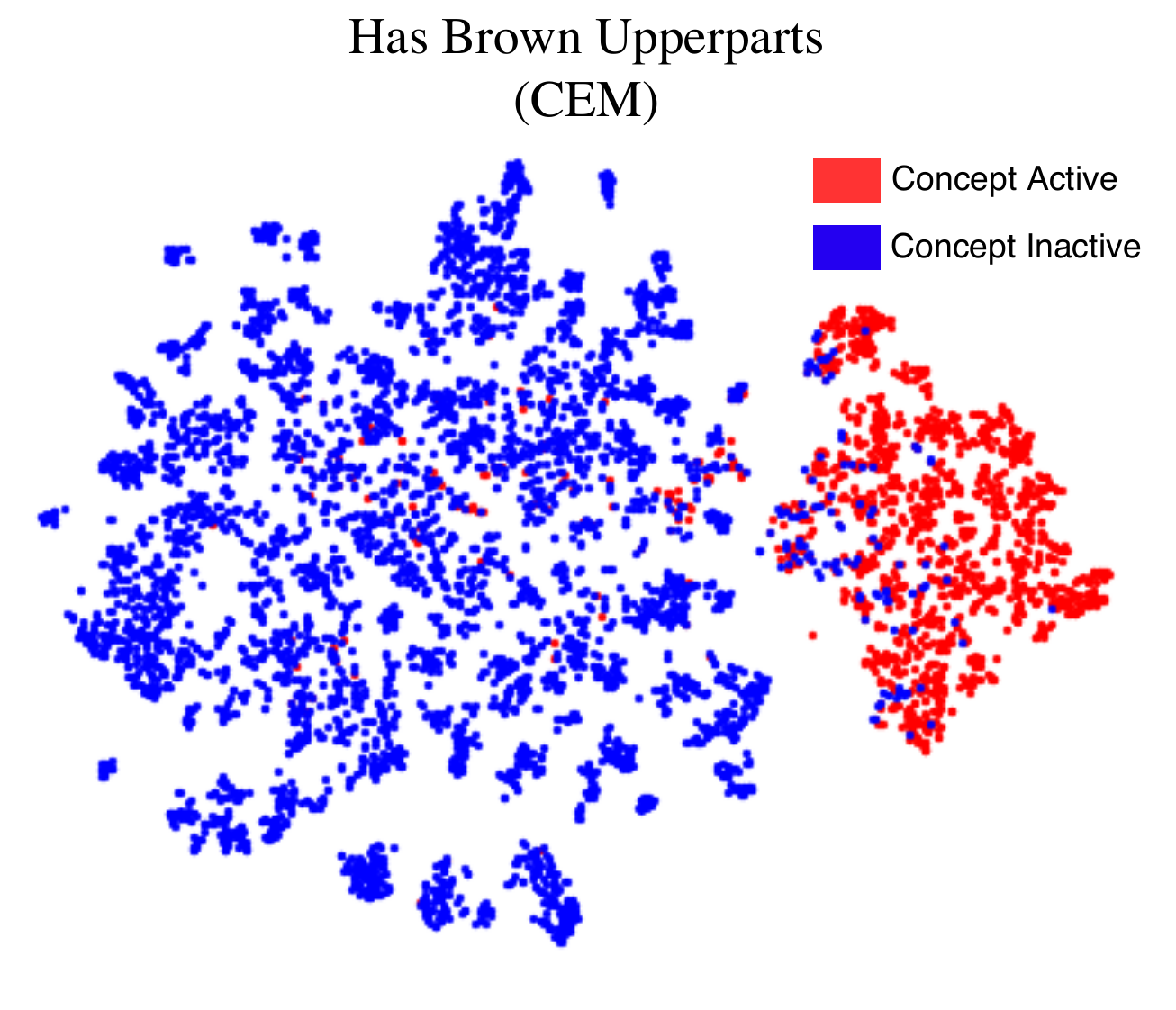}
        \includegraphics[width=\textwidth]{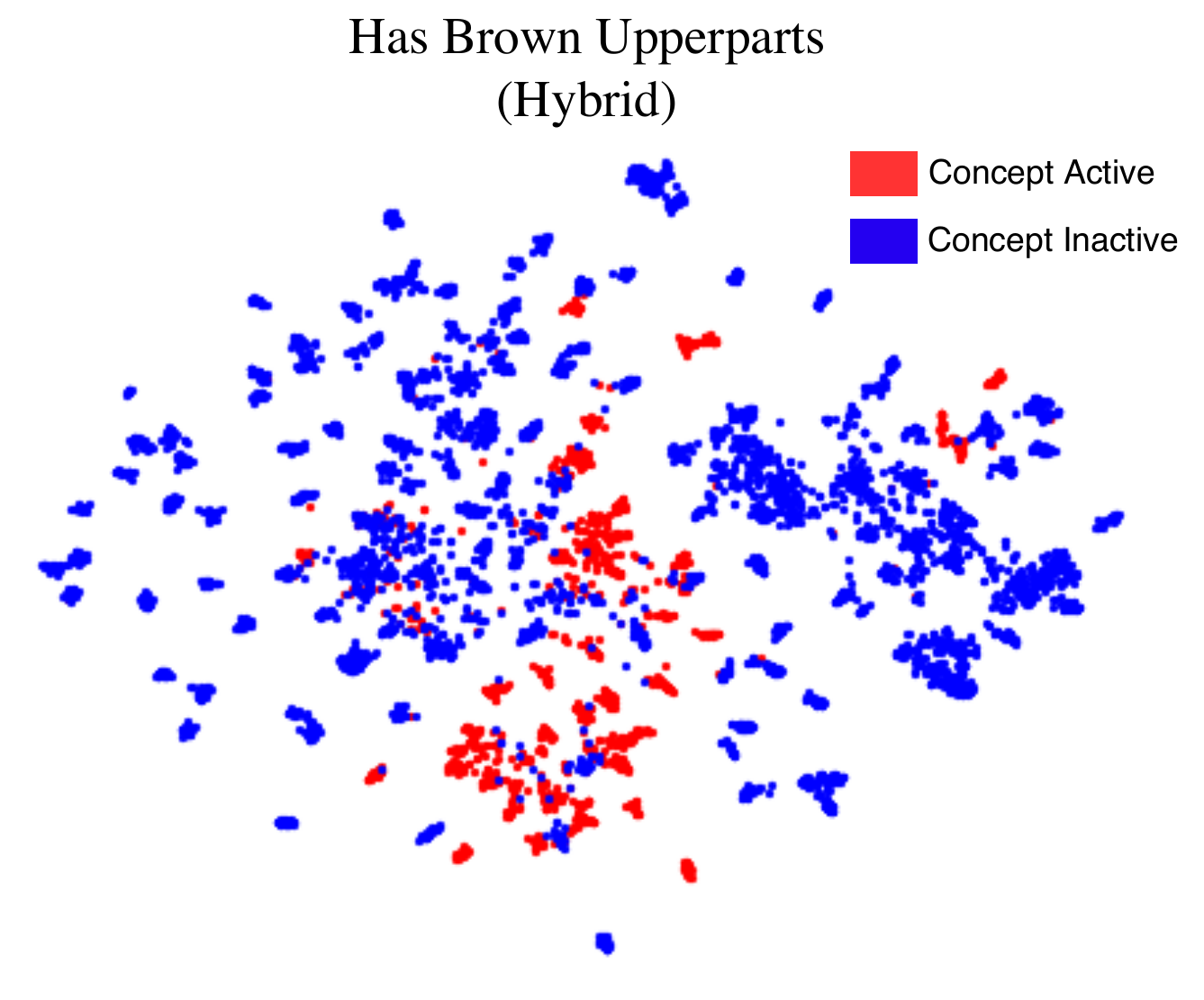}
        \subcaption{
        }
    \end{subfigure}
    \begin{subfigure}[b]{0.3\textwidth}
        \centering
        \includegraphics[width=\textwidth]{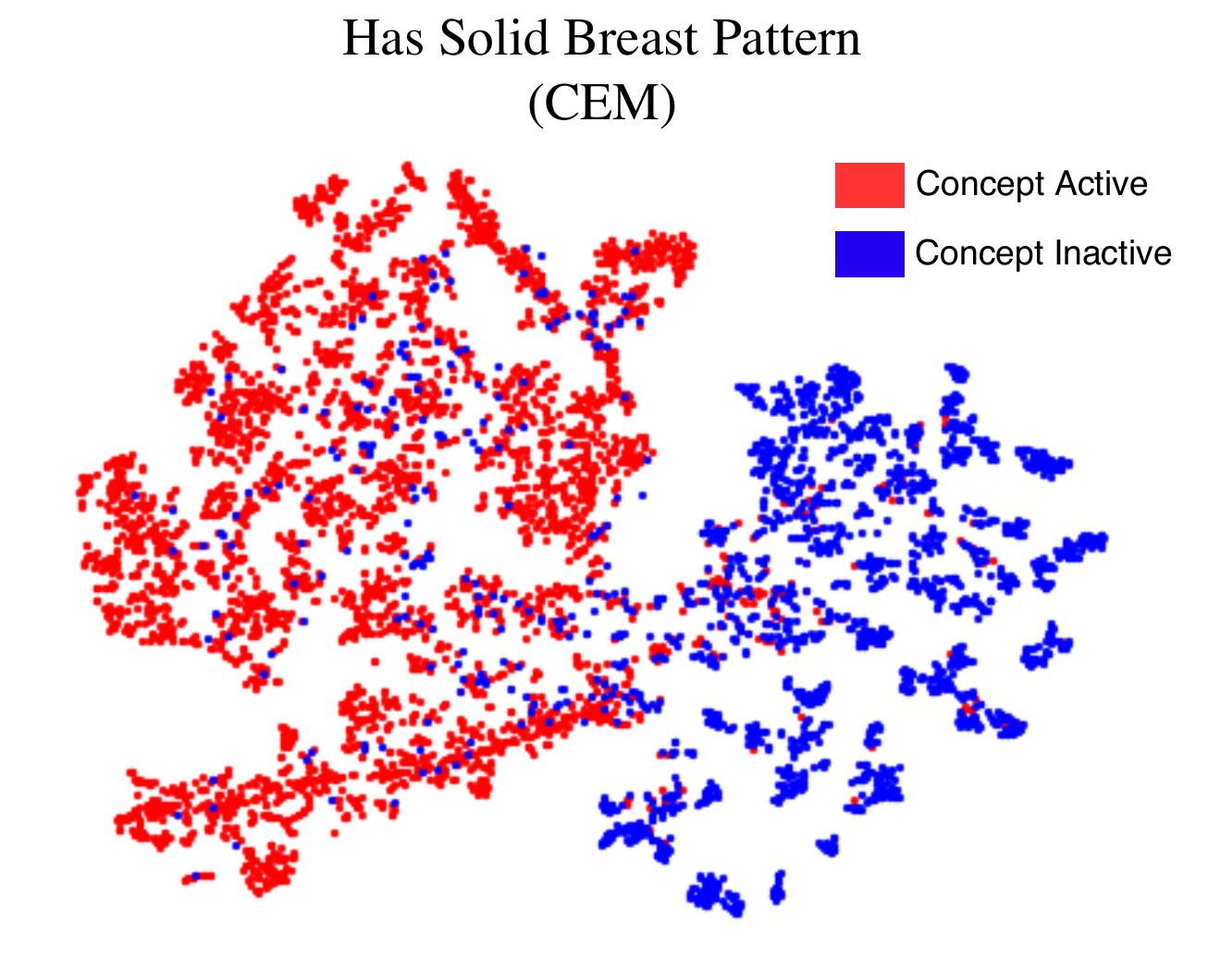}
        \includegraphics[width=\textwidth]{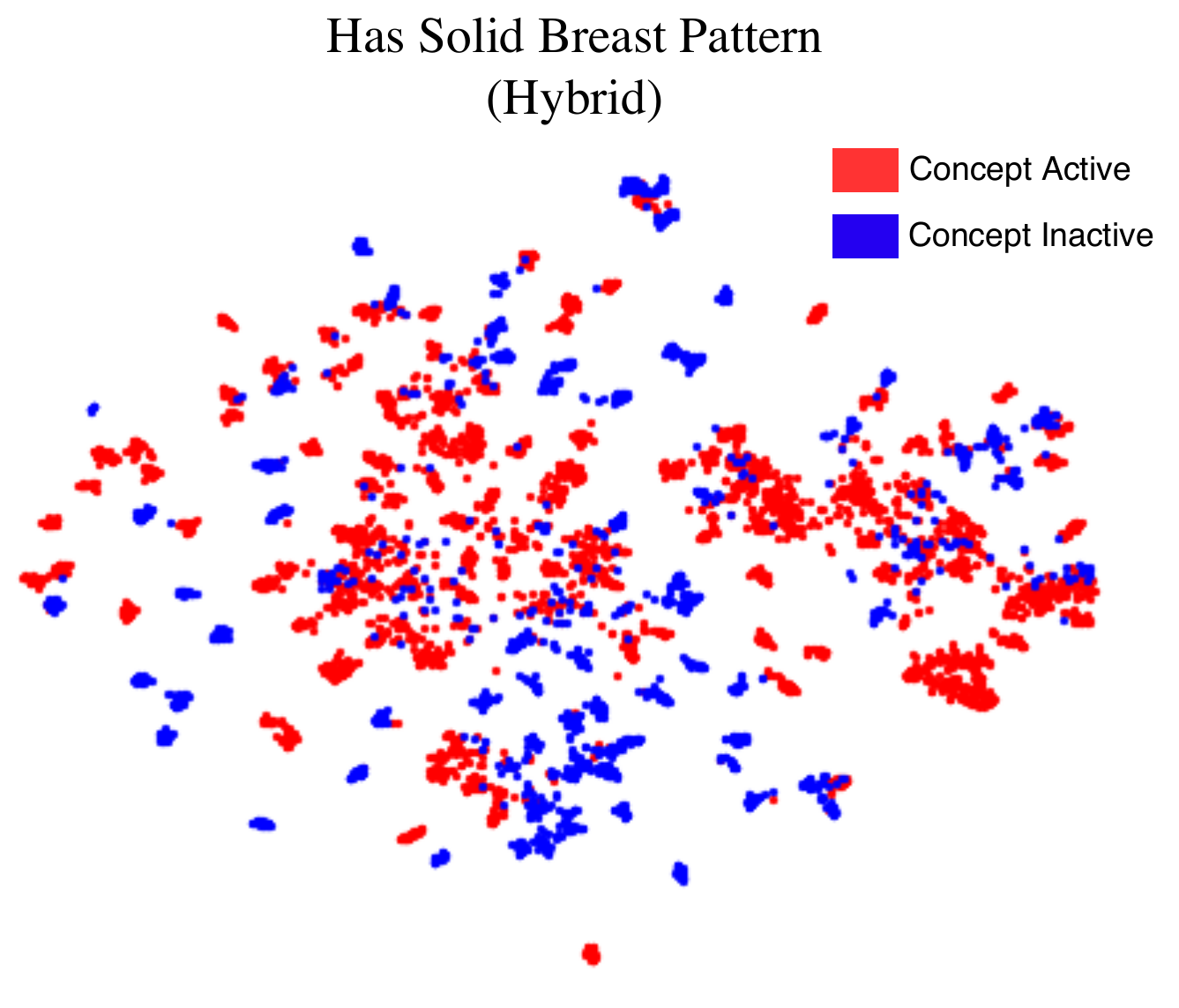}
        \subcaption{
        }
    \end{subfigure}
    \caption{t-SNE visualisations of CEM and Hybrid-CBM concept embeddings for concepts (a) ``has all purpose bill shape'', (b) ``has brown upperparts'', and (c) ``has solid breast pattern''. Each visualised test sample point is coloured red if the concept is active in that sample and blue otherwise. Concepts displayed in this figure were selected at random. All t-SNE plots are generated using a perplexity of $30$ and running the optimization for $1,500$ iterations.}
    \label{fig:appendix_tsne}
\end{figure}

In this section we show further qualitative results which highlight the same trends observed in Section~\ref{sec:experiments}. Specifically, we see via the t-SNE~\cite{van2008visualizing} plots shown in Figure~\ref{fig:appendix_tsne} that the concept representations learnt by Hybrid-CBMs are more visually entangled than those learnt by CEM. Notice that because in Hybrid-CBM we use $\mathbf{\hat{c}}_i = \hat{\mathbf{c}}_{[k:k + \gamma]}$ as the embedding learnt for concept $c_i$, all Hybrid-CBM t-SNE plots shown in Figure~\ref{fig:appendix_tsne} have a very similar arrangement and differ only in their coloring.

\begin{figure}[h!]
    \centering
    \begin{subfigure}[b]{0.3\textwidth}
        \centering
        \includegraphics[width=\textwidth]{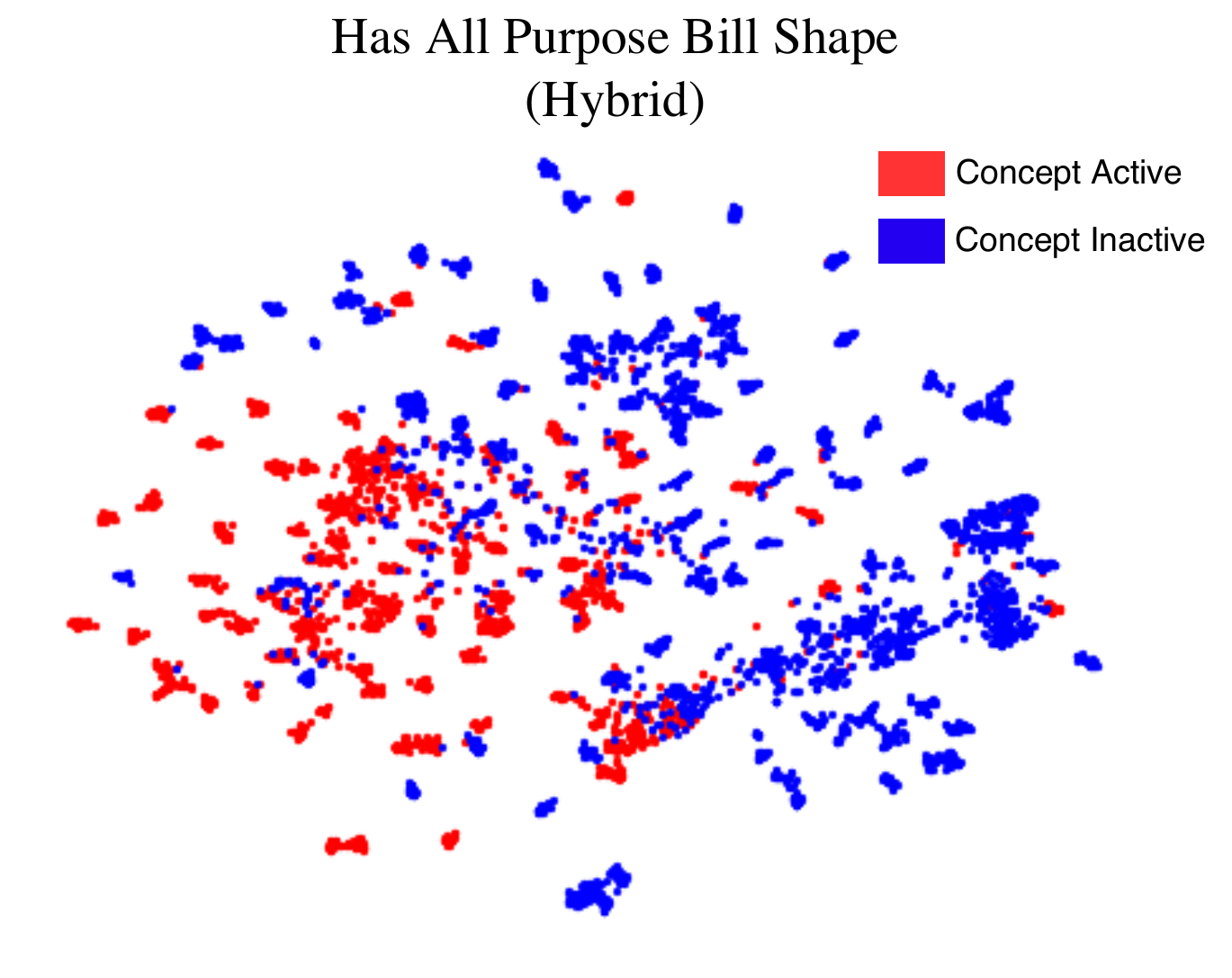}
        \subcaption{
        }
    \end{subfigure}
    \begin{subfigure}[b]{0.3\textwidth}
        \centering
        \includegraphics[width=\textwidth]{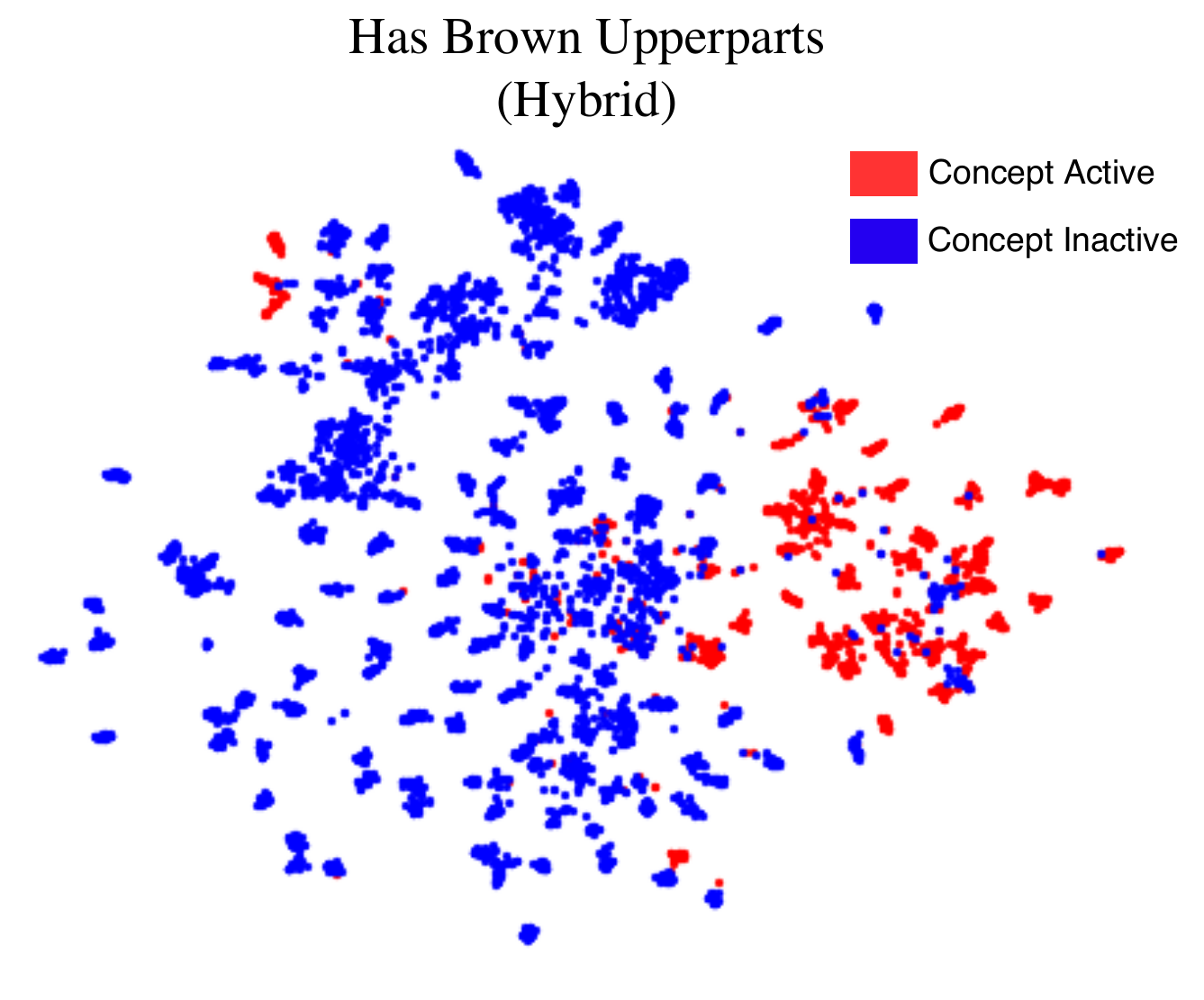}
        \subcaption{
        }
    \end{subfigure}
    \begin{subfigure}[b]{0.3\textwidth}
        \centering
        \includegraphics[width=\textwidth]{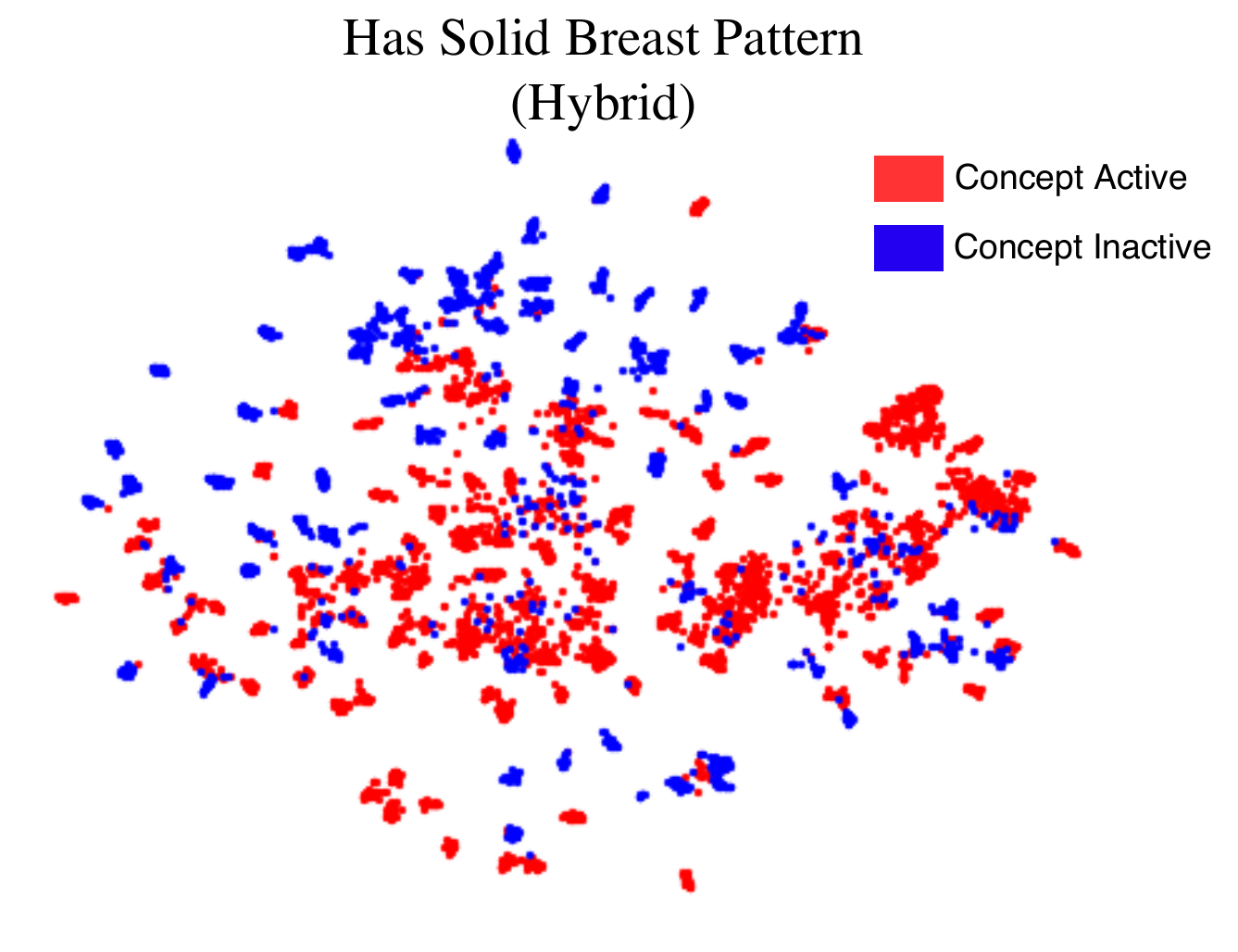}
        \subcaption{
        }
    \end{subfigure}
    \caption{t-SNE visualisations of Hybrid-CBM concept embeddings for concepts (a) ``has all purpose bill shape'', (b) ``has brown upperparts'', and (c) ``has solid breast pattern''. In contrast to the t-SNE plots shown in Figure~\ref{fig:appendix_tsne}, when producing these results we include the concept probability as part of the concept embedding learnt by Hybrid-CBM. All t-SNE plots are generated using a perplexity of $30$ and running the optimization for $1,500$ iterations.}
    \label{fig:appendix_tsne_with_probs}
\end{figure}

Moreover, Figure~\ref{fig:appendix_tsne_with_probs} shows that even when we include the concept probability as part of a concept's embedding in the Hybrid model (i.e., we let $\mathbf{\hat{c}}_i = [\hat{\mathbf{c}}_{[k:k + \gamma]}, \hat{\mathbf{c}}_{[i:(i + 1)]}]^T$ rather than $\mathbf{\hat{c}}_i = \hat{\mathbf{c}}_{[k:k + \gamma]}$ as before), we still observe similar entanglement within the latent space learnt for each concept in Hybrid-CBMs. This suggests that even when one includes a highly-discriminative feature, such as the probability of a concept being activated as part of the Hybrid-CBM's embeddings, the resulting representation is far from being easily separable w.r.t. its ground truth concept activation.

\begin{figure}[h!]
    \centering
    \begin{subfigure}[b]{0.32\textwidth}
        \centering
        \includegraphics[width=0.9\textwidth]{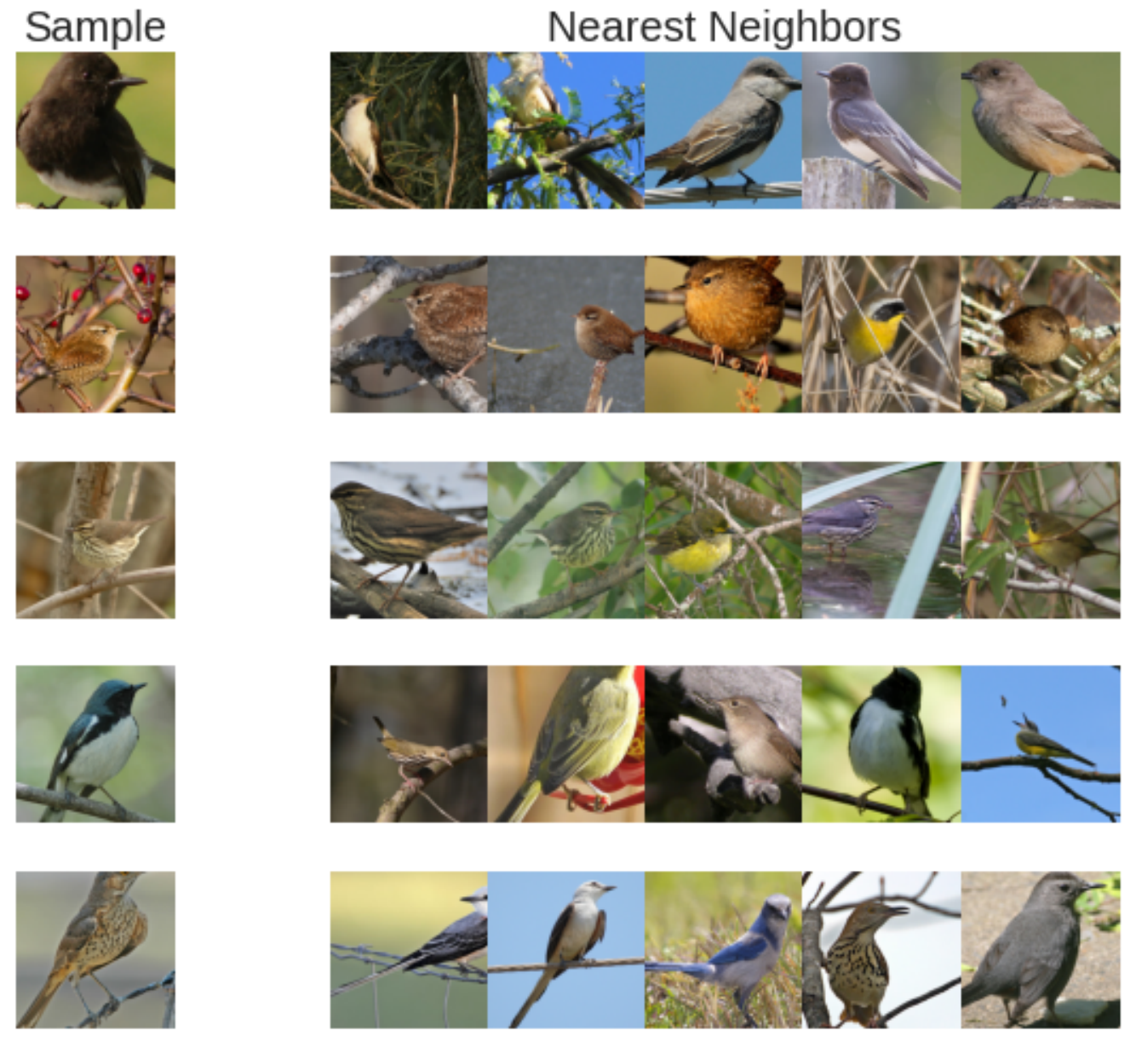}
        \subcaption{
        }
    \end{subfigure}
    \begin{subfigure}[b]{0.32\textwidth}
        \centering
        \includegraphics[width=0.9\textwidth]{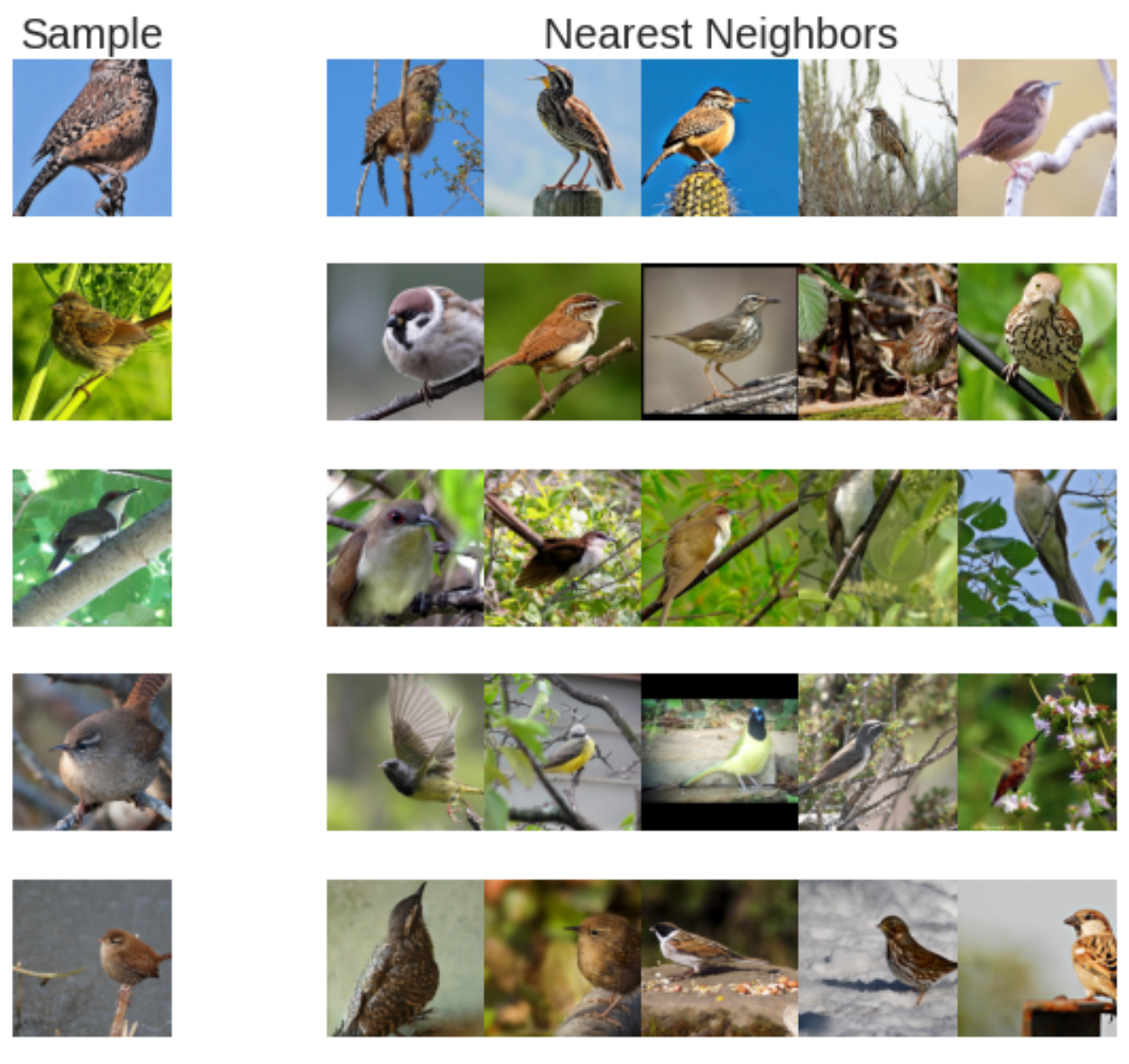}
        \subcaption{
        }
    \end{subfigure}
    \begin{subfigure}[b]{0.32\textwidth}
        \centering
        \includegraphics[width=0.9\textwidth]{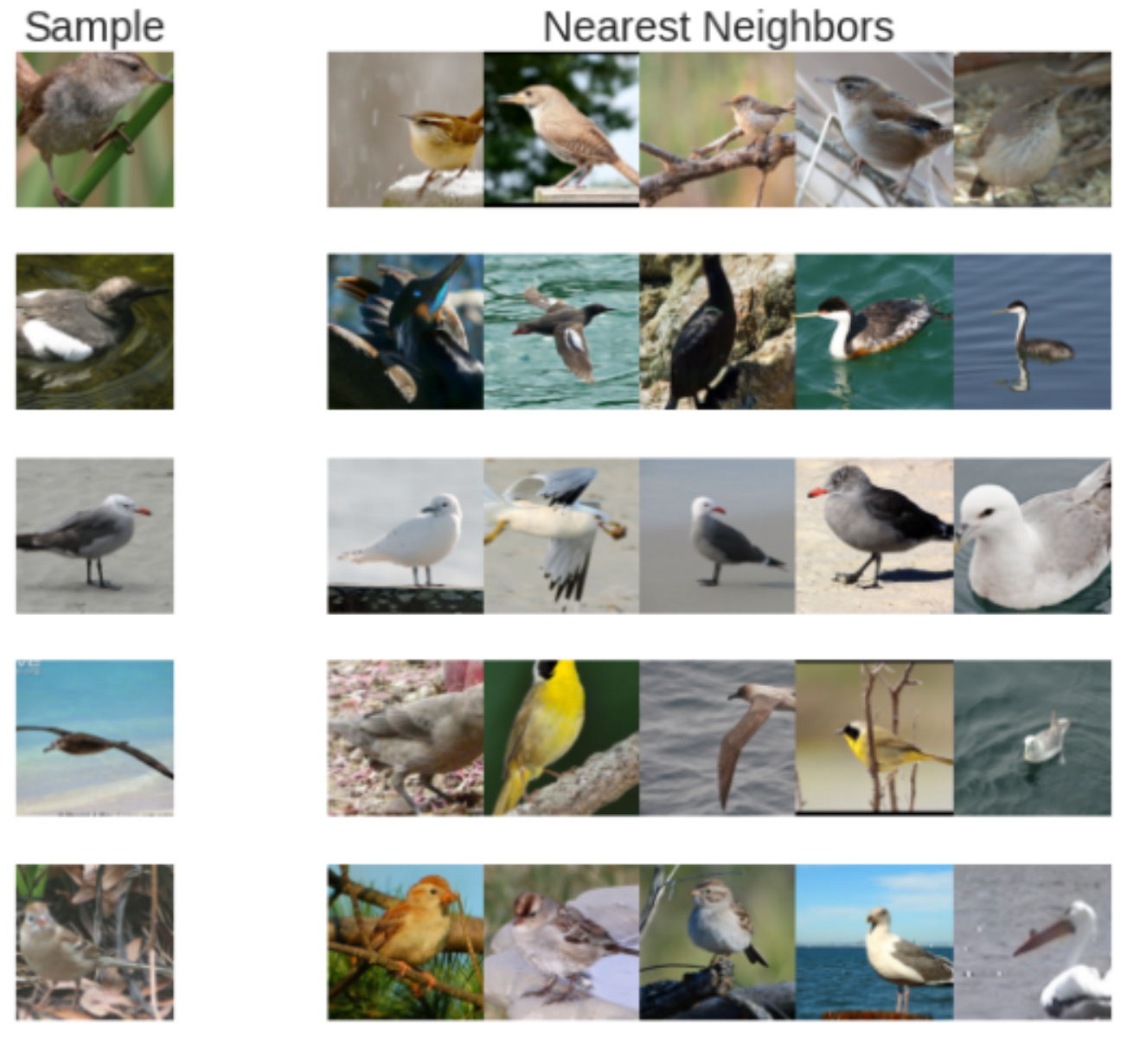}
        \subcaption{
        }
    \end{subfigure}
    \caption{Five nearest Euclidean neighbours to random test samples for concept embeddings (a) ``has all purpose bill shape'', (b) ``has brown upperparts'', and (c) ``has solid breast pattern''.}
    \label{fig:appendix_cem_nns}
\end{figure}

Finally, Figure~\ref{fig:appendix_cem_nns} shows that the coherency observed in Figure~\ref{fig:mixcem_nn} is seen across different learnt concept representations.

\subsection{Computational Cost of CEM}
\label{sec:appendix_computational_time}
\begin{figure}
    \centering
    \begin{subfigure}[h!]{0.8\textwidth}
        \includegraphics[width=\textwidth]{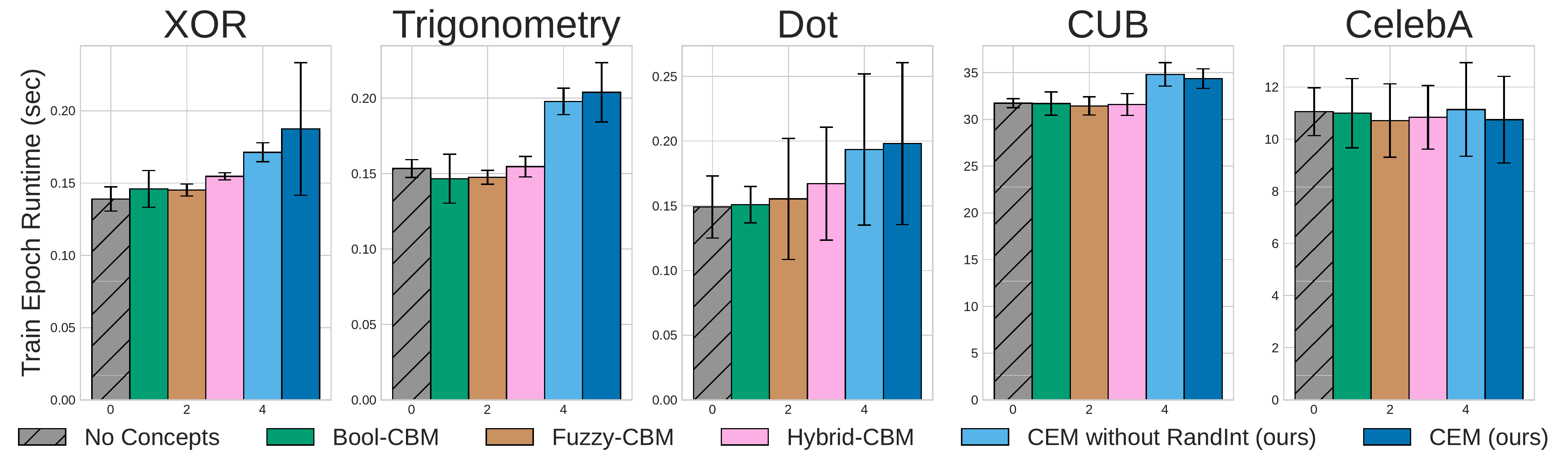}
        \subcaption{}
        \label{fig:epoch-runtime}
    \end{subfigure}
    \begin{subfigure}[b]{0.8\textwidth}
        \centering
        \includegraphics[width=\textwidth]{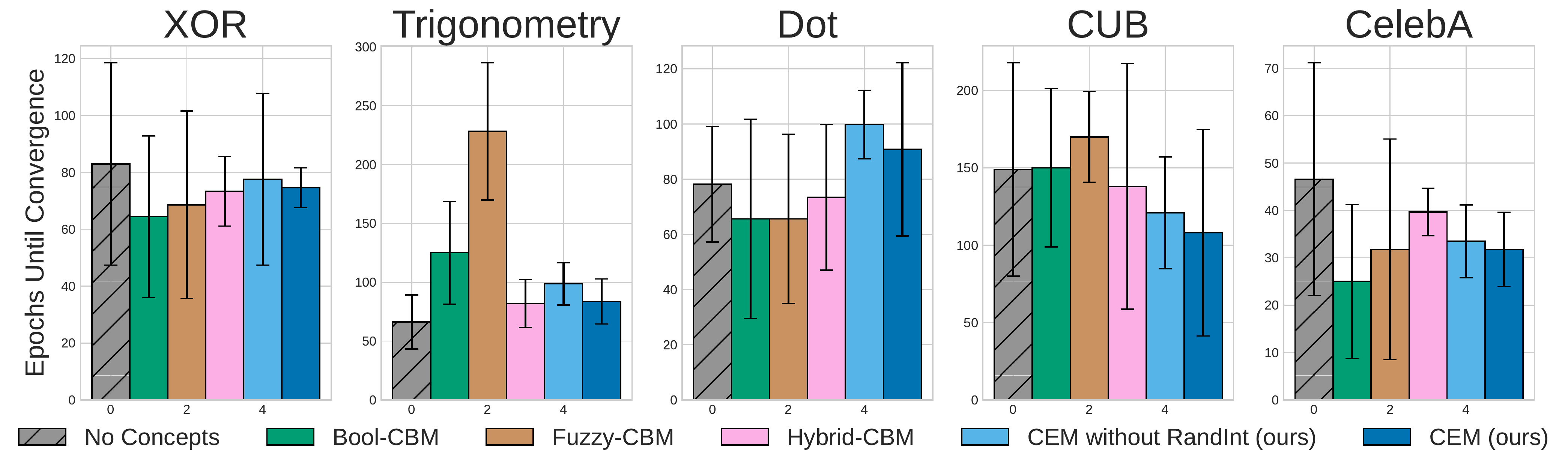}
        \subcaption{}
        \label{fig:epochs-conv}
    \end{subfigure}
    \caption{Computational cost of CEM compared to other baselines. (a) Average wall-clock runtime (in seconds) for one training epoch of each model. (b) Average number of training epochs performed until early stopping concluded the training run (recall we use a patience of 15 epochs).}
    \label{fig:train-time}
\end{figure}

\reb{
As mentioned in our Conclusion, CEM's use of an three linear layers (two for producing $\mathbf{\hat{c}}_i^+$ and $\mathbf{\hat{c}}_i^-$ and one for generating $\hat{p}_i$) leads to CEM requiring more FLOPs than vanilla CBMs per training epoch. Therefore, in this section we compare the computational cost of training CEM w.r.t. standard CBMs, by studying (i) the average runtime of one training epoch (Figure~\ref{fig:epoch-runtime}) and (ii) the average number of epochs taken for each method until convergence as dictated by our early stopping mechanism (Figure~\ref{fig:epochs-conv}). We observe that overall CEM does not incur in statistically significantly different training convergence times than other baselines. Similarly, as expected we see that a training step in CEM does require more FLOPs than vanilla CBMs (we empirically observe less than 10\% time increases in large datasets such as CUB and CelebA). Nevertheless, given its performance improvements showcased in Section~\ref{sec:experiments}, and its positive reaction to interventions, we believe that these small computational costs are justified.
}

\reb{Furthermore, we note that including RandInt in CEM does not significantly increase the training time in practice. This is due to the fact that its subroutine can be implemented using a simple multiplicative Bernoulli mask of the predictive concept probability vector.
}

\subsection{Effect of RandInt in standard CBMs}
\label{sec:appendix_randint_cbms}
\begin{figure}[!h]
    \centering
    \begin{subfigure}[b]{0.85\textwidth}
        \includegraphics[width=\textwidth]{fig/interventions_complete_with_adversarial.pdf}
        \subcaption{}
    \end{subfigure}
    \begin{subfigure}[b]{0.85\textwidth}
        \includegraphics[width=\textwidth]{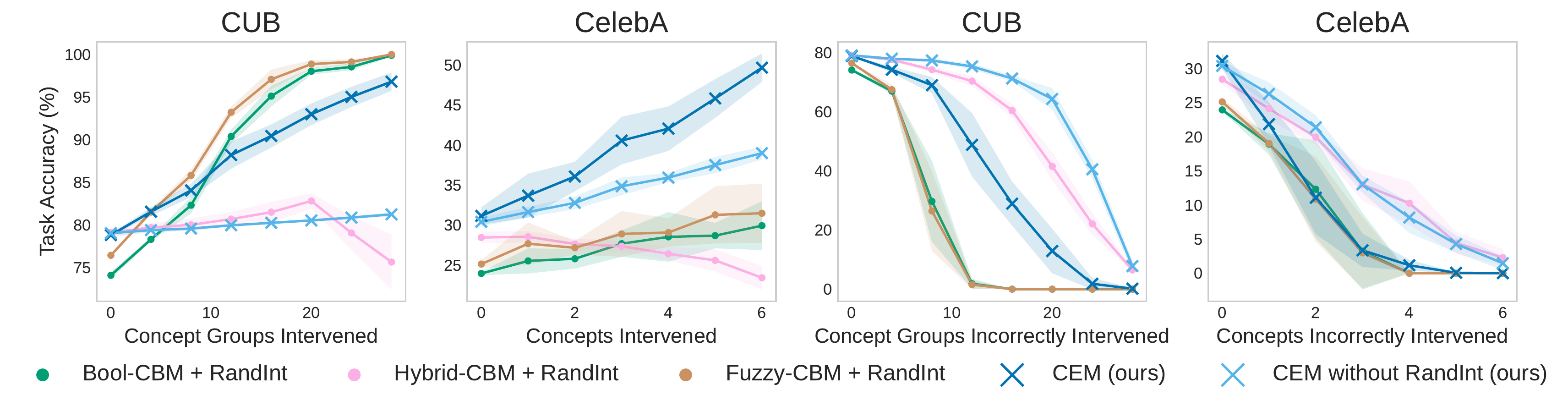}
        \subcaption{}
    \end{subfigure}
    \caption{Task accuracy after interventions with and without RandInt for all methods. (a) Task accuracy after both ``correct'' and ``incorrect'' interventions for models trained without RandInt. (b) Task accuracy after both ``correct'' and ``incorrect'' interventions for models trained with RandInt.}
    \label{fig:randint-cbms}
\end{figure}

\reb{
RandInt is a form of regularization that we specifically designed to applicable to CEM's use of a positive and negative concept embedding. Its purpose is to incentivize each embedding to be better aligned with the ground truth semantics it represents so that their use in interventions is more effective. Nevertheless, as it is formulated in Section~\ref{sec:randint}, it is possible to apply it to other kinds of CBMs (e.g., Fuzzy and Hybrid CBMs). When applied to other kinds of models, however, it may not have the intended effect. For example, in vanilla CBMs where there is no extra capacity in the bottleneck, RandInt will behave in a similar way to a dropout regularizer and may instead force the label predictor to depend less on a specific concept activation when the concepts are an incomplete description of the task (therefore leading to possibly worse responses to concept interventions). Notice that this does not happen in CEM as during training RandInt still allows gradients to flow and update the weights that generate the “correct” embedding, letting the model modify this embedding so that it is aligned with its intended semantics. On the other hand, if the concepts are a complete description of the downstream task, then, as $p_\text{int}$ approaches 1, we expect RandInt's use in a CBM to behave similarly to how a independently-trained CBM behaves (where the concept encoder and label predictor models are trained separately). This means that, as shown in~\cite{koh2020concept}, it may lead to some improvements in how effective interventions are. To verify this, and for a fair comparison across methods, we train all CBM baselines with our RandInt regularizer ($p_\text{int} = 0.25$ as in the rest of experiments). As hypothesized, we observe in Figure~\ref{fig:randint-cbms} that RandInt seems in fact to hurt the performance of standard CBMs in concept-incomplete tasks (e.g., CelebA) while it adds small performance improvements in concept-complete tasks (e.g., CUB).
More importantly, however, notice that our main result of our intervention results in Section~\ref{sec:int_results} still hold: CEM still significantly outperforms Hybrid-CBMs, its closest competitor, even when the Hybrid model is trained with RandInt.
}

\subsection{Intervention Experiment Details}
\label{sec:appendix_intervention}

\paragraph{Setup} For our intervention results discussed in Section~\ref{sec:experiments}, for each method we train 5 different models using different random seeds. Then, when intervening on a model $\mathcal{M}$ by correcting $d$ of its concepts at test-time, we select the same random subset of $d$ concepts we will intervene on for all models trained with the same initial seed as $\mathcal{M}$. Given that several CUB concept annotations are mutually exclusive (e.g., ``has white wings'' and ``has brown wings''), following~\citep{koh2020concept} when intervening in models trained in this task we jointly set groups of mutually exclusive concepts to their ground truth values. This results in a total of $28$ groups of mutually exclusive concepts in CUB which we intervene on.

\paragraph{Exploring effects of different training procedures in CBM interventions}

Previous work by~\citet{koh2020concept} suggests that CBMs trained sequentially (where the concept encoder is trained first and then frozen when training the label predictor) or independently (where the concept encoder and label predictor are trained independently of each other and then composed at the end to produce a CBM) can sometimes outperform jointly trained CBMs when expert interventions are introduced. In this section we explore whether the results shown in Figure~\ref{fig:interventions} would differ if one compares our model against sequentially and independently trained Fuzzy-CBMs.

\begin{figure}[!h]
    \centering
    \includegraphics[width=0.9\textwidth]{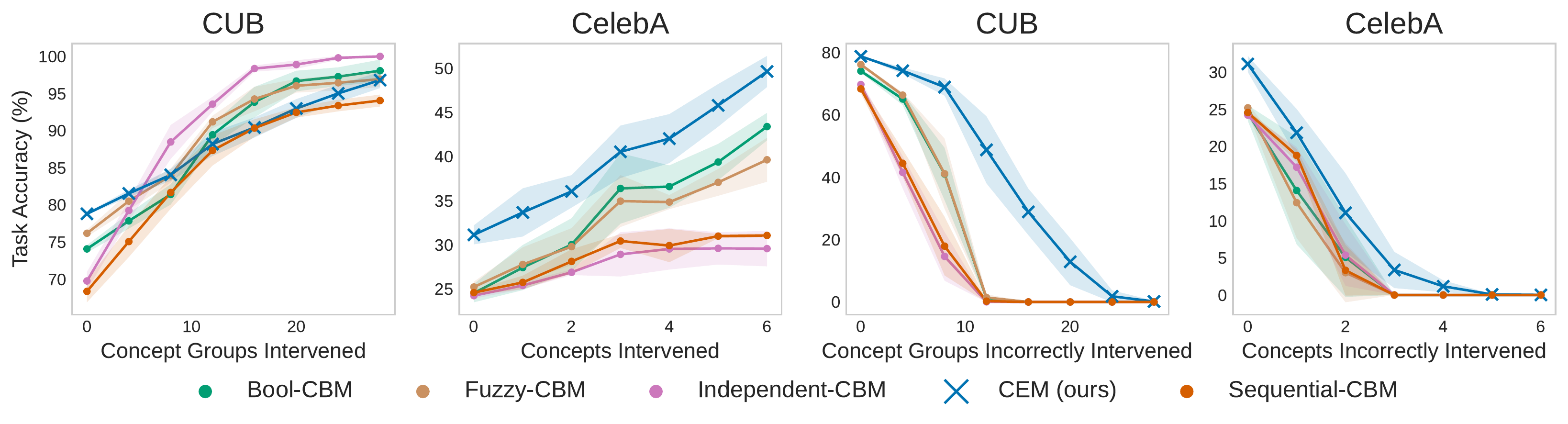}
    \caption{Effects of performing positive random concept interventions (left and center left) and incorrect random interventions (center right and right) for different training regimes for CBMs (Joint, Sequential, and Independent). For clarity, Hybrid is not included in this plot (see Figure~\ref{fig:interventions} for those results).}
    \label{fig:appendix_interventions_train_regs}
\end{figure}

Figure~\ref{fig:appendix_interventions_train_regs} shows how CEMs react to interventions compared to sequentially and independently trained CBMs. Notice that the observed trends in these results are not so different than those seen when comparing CEMs against jointly-trained CBMs: in concept completeness (e.g., CUB), Fuzzy-CBMs (with the exception of Sequential-CBMs which seem to underperform) tend to react better to correct interventions than CEM but can quickly drop their performance if these interventions are not correct. In stark contrast, however, in concept-incomplete settings such as in CelebA, we see that Sequential and Independent CBMs experience mild performance improvements when correct interventions are performed, leading to CEMs outperforming these models by a large margin. These results suggest that our observations in Section~\ref{sec:experiments} hold even if one changes the training process for a Fuzzy-CBM and highlight that CEMs are the only models in our evaluation capable of maintain high performance both in concept-complete and concept-incomplete settings.

\subsection{Embedding Size Ablation Study}
\label{sec:appendix_emb_size_ablation}
\begin{figure}[h!]
    \centering
    \begin{subfigure}[b]{0.55\textwidth}
        \includegraphics[width=\textwidth]{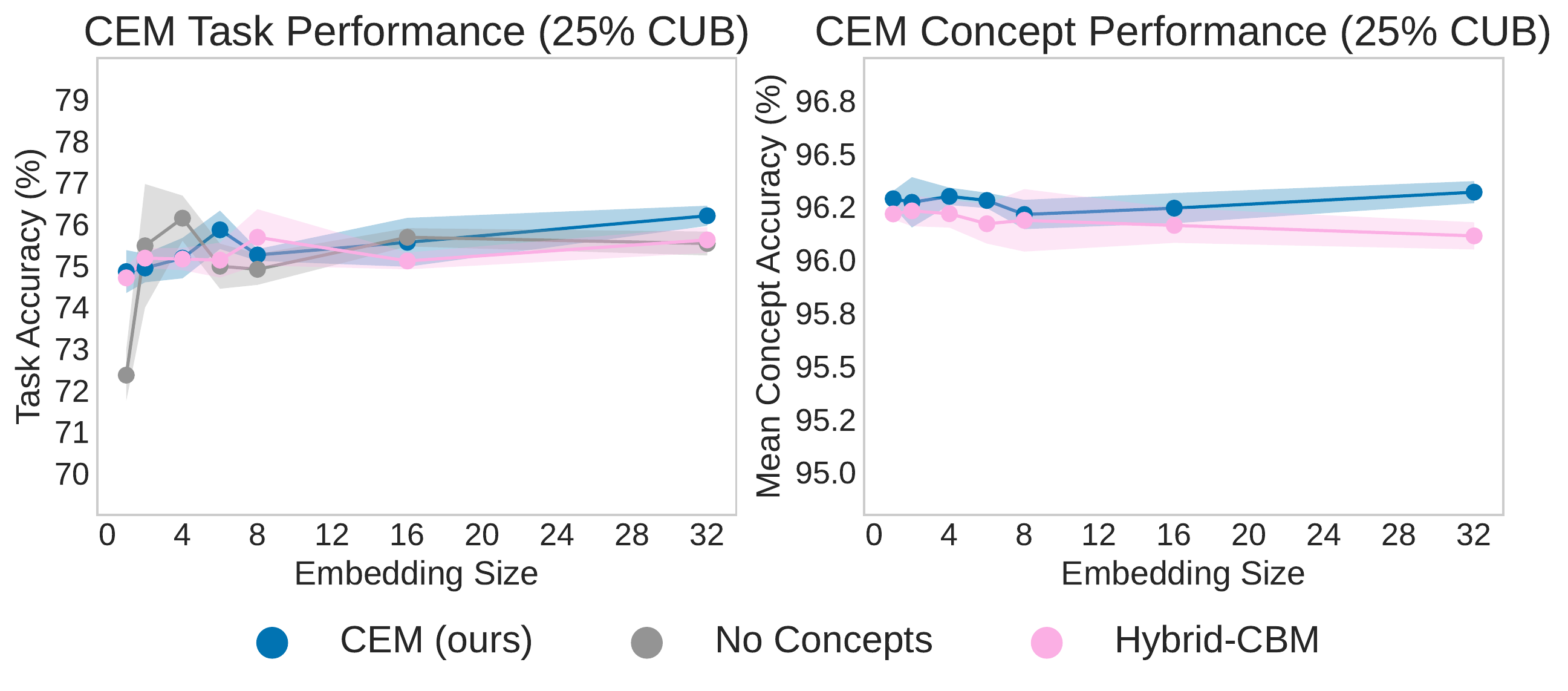}
        \subcaption{}
    \end{subfigure}
    \begin{subfigure}[b]{0.4\textwidth}
        \includegraphics[width=\textwidth]{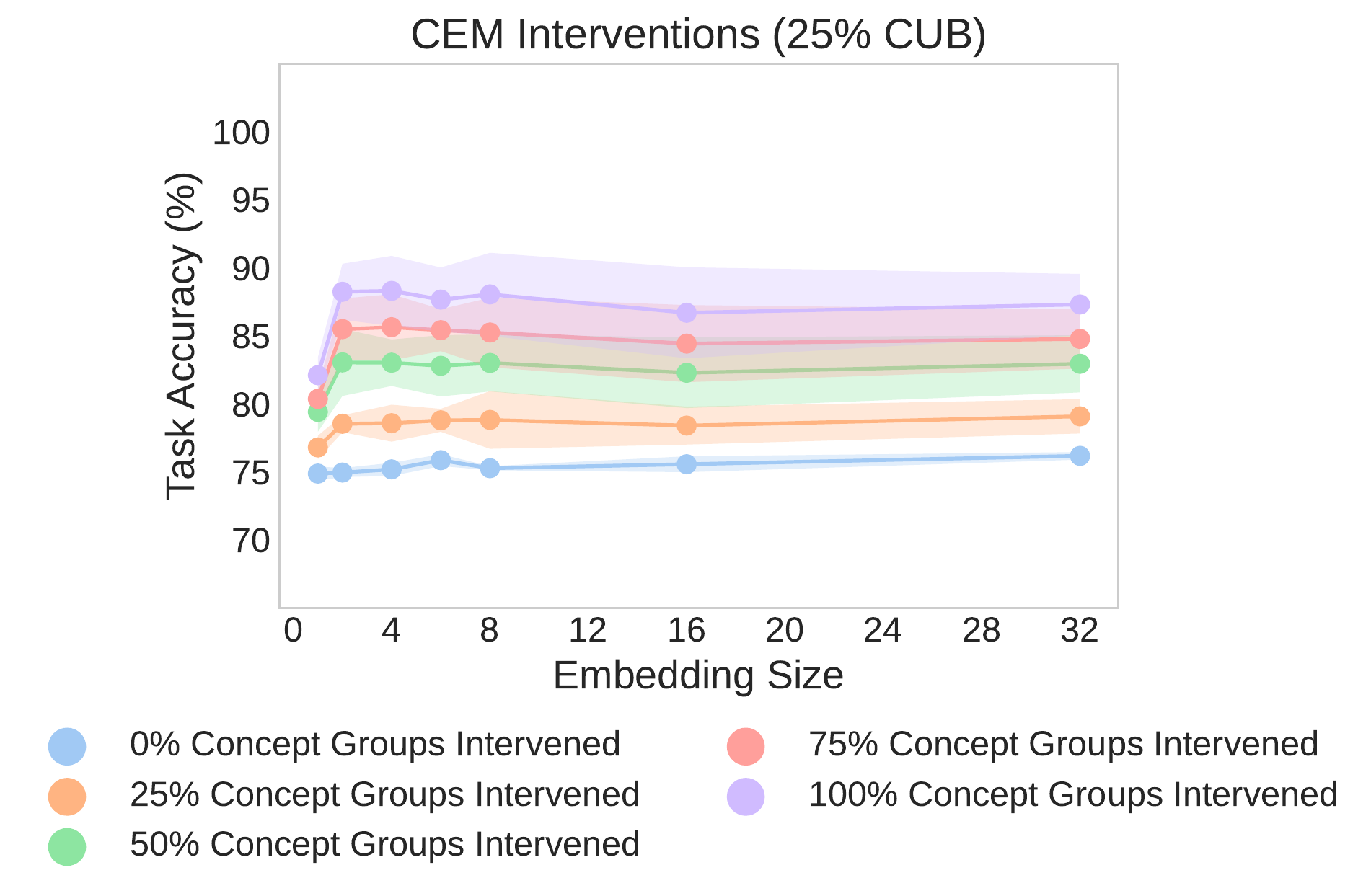}
        \subcaption{}
    \end{subfigure}
    \caption{Ablation study for $m$ in CUB when only 25\% of its concept annotations are used during training. (a) Task and concept validation accuracy of CEMs with different embedding sizes. For comparison, we include Hybrid-CBMs and end-to-end black box models with equal bottleneck capacity as their CEM counterpart for each value of $m$. (b) Task validation accuracy when intervening on an increasing number of concept groups for CEMs with different embedding sizes.}
    \label{fig:appendix_emb_size_ablation_cub}
\end{figure}

\begin{figure}[h!]
    \centering
    \begin{subfigure}[b]{0.55\textwidth}
        \includegraphics[width=\textwidth]{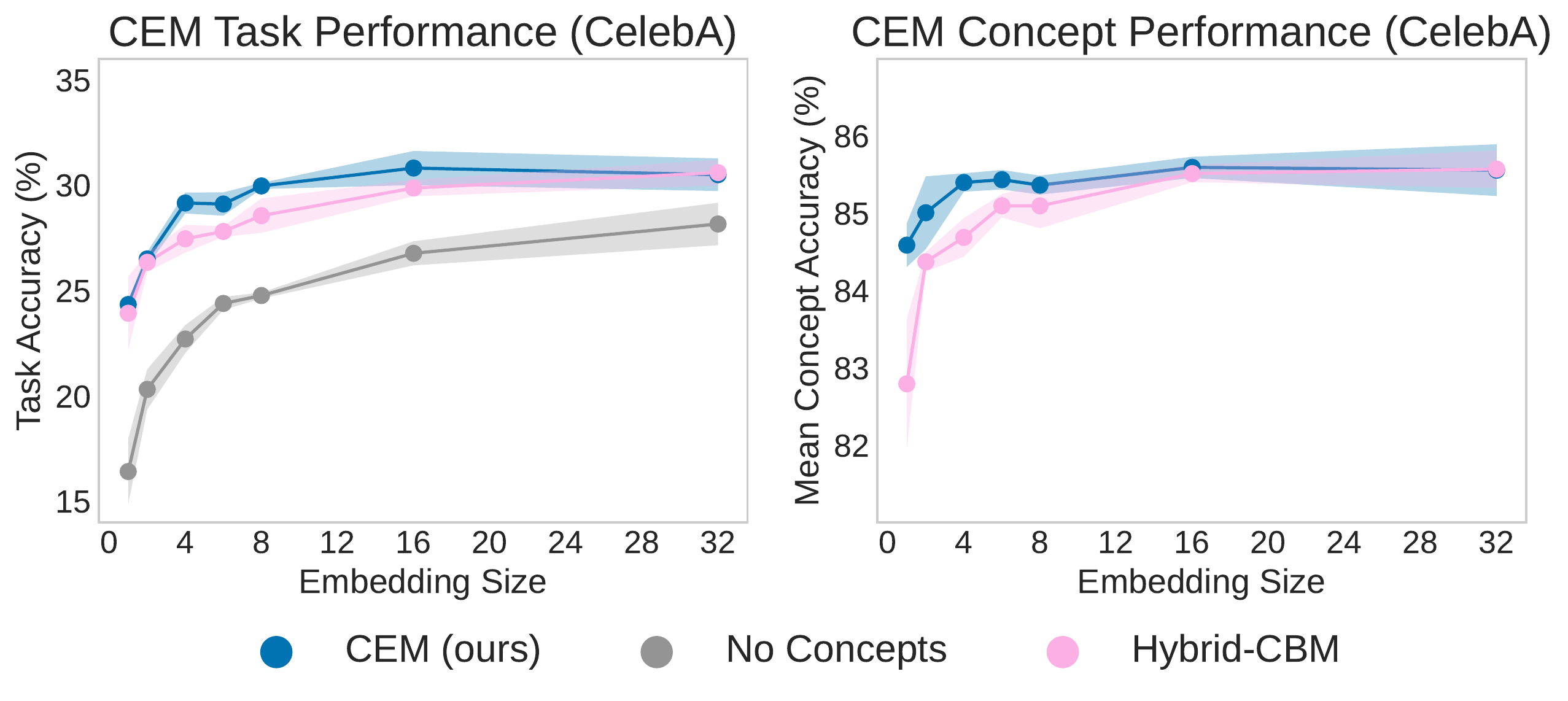}
        \subcaption{}
    \end{subfigure}
    \begin{subfigure}[b]{0.4\textwidth}
        \includegraphics[width=0.9\textwidth]{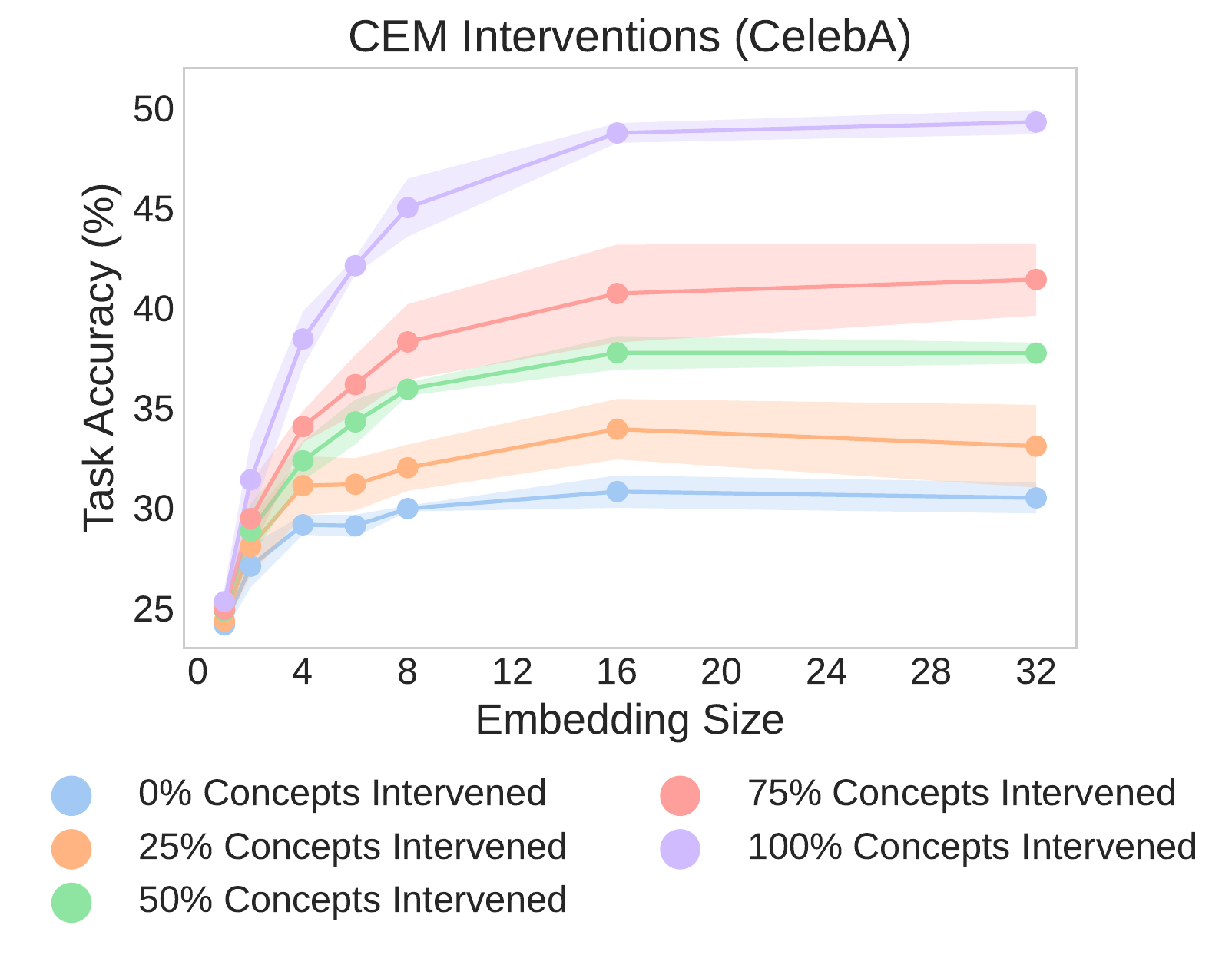}
        \subcaption{}
    \end{subfigure}
    \caption{Ablation study for $m$ in CelebA. (a) Task and concept validation accuracy of CEMs with different embedding sizes. (b) Task validation accuracy when intervening on an increasing number of concepts for CEMs with different embedding sizes.}
    \label{fig:appendix_emb_size_ablation_celeba}
\end{figure}

In this section we explore the effects of the embedding size $m$ in CEMs and compare their performance as $m$ varies against that of Hybrid-CBMs and end-to-end black box models with equal capacity. For this, we train CEMs, Hybrid-CBMs, and end-to-end black box models on CUB (with only 25\% of its concept annotations being selected) and CelebA using the same architectures and training configurations as described in Appendix~\ref{sec:appendix_architectures}. We chose to reduce the number of concept annotations in CUB to better study how our model behaves when the raw number of activations in its bottleneck (which is equal to $(k \cdot m)$ in CEMs) is severely constrained. We show our results in Figure~\ref{fig:appendix_emb_size_ablation_cub} and Figure~\ref{fig:appendix_emb_size_ablation_celeba}.

Our study shows that, for both tasks, after enough capacity is provided to CEMs (which for our particular datasets seems to be around 8-16 activations per embedding), our models are able to perform better or competitively against end-to-end black box models and Hybrid-CBMs. In particular, we see that with the exception of very small embedding sizes, CEM tends to outperform Hybrid-CBM models with equal capacity, suggesting that introducing a fully supervised bottleneck can in fact help in both task and mean concept performance. Similarly, we see that with the exception of when the embedding size is $m = 2$ in CUB, CEMs are able to perform equally as well or better than end-to-end black box architectures with equal capacity. Furthermore, notice that even in the case where end-to-end black box models outperform CEM (as in $m=2$ for CUB), the difference in task accuracy is less than $1.5\%$, a hit which may not be detrimental if one takes into account the fact that CEM produces highly-accurate concept-based explanations and it is able to significantly surpass the performance of end-to-end black box model if interventions in its concept bottleneck are allowed. Finally, we similarly see for both tasks that interventions have similar effects on models after a moderately sized embedding is used, therefore suggesting there is no benefit in increasing the embedding size significantly if one is interested in interventions. These two studies suggest that unless the embedding size is drastically constrained (e.g., $m \leq 4$), CEM's performance is stable with respect to the embedding size used, aiding with hyperparameter selection and allowing CEMs to be more easily integrated into other architecture designs.

\subsection{Code, Licences, and Resources} \label{sec:appendix_code}

\paragraph{Libraries} For our experiments, we implemented all baselines and methods in Python 3.7 and relied upon open-source libraries such as PyTorch 1.11~\citep{paszke2019pytorch} (BSD license) and Skelearn~\citep{ pedregosa2011scikit} (BSD license). To produce the plots seen in this paper, we made use of Matplotlib  3.5 (BSD license). We have released all of the code required to recreate our experiments in an MIT-licensed public repository\footnote{Code can be found at \href{https://github.com/mateoespinosa/cem/}{https://github.com/mateoespinosa/cem/}}.

\paragraph{Resources} All of our experiments were run on a private machine with 8 Intel(R) Xeon(R) Gold 5218 CPUs (2.30GHz), 64GB of RAM, and 2 Quadro RTX 8000 Nvidia GPUs. We estimate that approximately 240-GPU hours were required to complete all of our experiments.




\end{document}